\documentclass[12pt]{article}
\usepackage{amsmath}
\usepackage{graphicx,psfrag,epsf}
\usepackage{enumerate}
\usepackage{natbib, color}

\usepackage[ruled,vlined]{algorithm2e}
\usepackage{mathtools}
\usepackage{diagbox}
\usepackage{color, colortbl, xcolor}
\usepackage{multirow}
\usepackage{algorithmicx}
\usepackage{algpseudocode}
\usepackage{amsmath,amsfonts,bm}
\usepackage{amssymb}
\usepackage{amsthm}
\usepackage{graphics}
\usepackage{epsfig}

\usepackage{float}
\usepackage{subcaption}

\usepackage{xr}
\DeclareMathOperator*{\argmax}{argmax}
\usepackage{mathtools}

\theoremstyle{plain}
\newtheorem{thm}{Theorem}[section]

\theoremstyle{definition}

\theoremstyle{remark}

\begin{document}
\centerline{}
\vskip 3mm

\noindent Determination of class-specific variables in nonparametric multiple-class classification
\vskip 3mm

\vskip 5mm
\noindent Wan-Ping Nicole Chen

\noindent Institute of Statistical Science

\noindent Academia Sinica

\noindent Taipei, Taiwan 11529

\noindent wpchen@stat.sinica.edu.tw

\vskip 5mm
\noindent Yuan-chin Ivan Chang

\noindent Institute of Statistical Science

\noindent Academia Sinica

\noindent Taipei, Taiwan 11529

\noindent ycchang@stat.sinica.edu.tw

\vskip 3mm

\noindent ABSTRACT
{
As technology advanced, collecting data via automatic collection devices become popular,
thus we commonly face data sets  with lengthy variables, especially when
these data sets are collected without specific research goals beforehand.
It has been pointed out in the literature that the difficulty of high-dimensional classification problems is intrinsically caused by too many noise variables useless for reducing classification error, which  offer less benefits for decision-making, and increase complexity, and confusion in model-interpretation.
A good variable selection strategy is therefore a must for using such kinds of data well; especially when we expect to use their results for the succeeding applications/studies, where the model-interpretation ability is essential.
hus, the conventional classification measures, such as accuracy, sensitivity, precision, cannot be the only performance tasks.
In this paper, we propose a probability-based nonparametric multiple-class classification method, and 
integrate it with the ability of identifying high impact variables for individual class such that we can have more information about its classification rule and the character of each class as well.
The proposed method can have its prediction power approximately equal to that of the Bayes rule, and still retains the ability of ``model-interpretation.''
We report the asymptotic properties of the proposed method, and use both synthesized and real data sets to illustrate its properties under different classification situations.
We also separately discuss the variable identification, and training sample size determination, and summarize those procedures as algorithms such that users can easily implement them with different computing languages.
}
%

%
   
\section{Introduction}
In multiple-class classification problems, the relative position among classes can be very complicated within the variable-space, 
the situation becomes worse for data sets with lengthy variables.
Modern machine learning methods commonly outperform the conventional statistical classification models in these situations. 
These methods, such as the support vector machine (SVM), may have good prediction power;
however, the classification rules of them are difficult to be explained.
There are also some methods build multiple-class classification rules baed on the divide-and-conquer concept via a bunch of binary classifiers, and thus 
rely on certain voting schemes, which make model-interpretation more difficult \citep[see also][]{tomczak2015}.  
Some hybrid methods, which depend on voting-like schemes, such as random forest methods, will also suffer from model-interpretation difficulty. 
These nonparametric types of methods are preferred due to their flexibility, especially for problems with complicated data, such as image, and sound, where conventional parametric models are inferior.
However, their classification rules are less informative for the follow-up applications/studies, where the further details of classification rules, or  ``model interpretation'' property is essential in addition to the conventional performance measures such as accuracy, precision, and so on.
Thus, the interpretation ability of classifiers catches a lot of attention \citep{Yang2010}, and  also motivates this study. 

For conventional classifier construction processes, when a data set has a large number of variables, we aim to find a short set of informative variables to build ``separating'' boundaries among classes for the ease of classification rules interpretation. 
This task is crucial even in binary classification problems \citep[see][]{Liu2008, mechbal2015, Zhang2017}, 
and it becomes more difficult to achieve in multiple-class classification problems due to the complications among classes.
Moreover, if all the conditional probabilities, given each class label, of a sample are known, we can assign this sample to the class where its largest conditional probability.   
This classification rule is called the Bayes rule, which has the smallest error rate \citep[see][]{ryzin1966}. 
In \citet{ryzin1966}, he also showed that if we accurately estimate each class density, then the classification error rate approaches to that of the Bayes rule.
Since then, the probability-based classification rules have been extensively discussed in the literature
\citep{ 
DBLP:journals/tit/DevroyeW76, 
DBLP:journals/prl/GreblickiP83, 
Har-Peled:2002, 
John95estimatingcontinuous}. 
Although there are many different probability-based methods reported, the model interpretation ability is usually out of their scopes \citep{Ancukiewicz1998, hall2005, lugosi1996, Kobos2011, 796369}. 
However, to find only one  variable set for all classes in a multiple-class classification problem makes the estimation of the density of each class very difficult, since all classes have their own characters, and may not be described via the same set of variables.
Thus, how to retain prediction power, while to identify proper variables of each class for improving the model-interpretation ability for multiple-class classifiers is our goal in this paper.

\cite{shao2011} and many other researchers have pointed out that a classification rule could do as bad as random guessing without considering the sparsity condition, and complex structure in high-dimensional data sets.
\cite{fan2008} state that the difficulty of high-dimensional classification problems is intrinsically caused by too many noise variables useless for reducing classification error, which  offer less benefits for decision-making, and increase complexity, and confusion in model-interpretation. 
Moreover, they also pointed out that the ``important variables'' for estimating the conditional densities for individual classes can be different, 
and this fact makes the model-interpretation of the probability-based methods even more difficult for multiple-class classification problems.
Thus, identifying the individual high impact variable set of each class plays a crucial role in both classification performance, and model-interpretation, especially
when the follow-up research requires the details of the classification rules instead of just the predicted labels.

We study multiple-class classification problems under sparsity condition, where each sample belongs to only one class.
Our goal is to assign new sample to  the most appropriate class, 
and learn about ``why this sample should be assigned to a particular class'' through the identified variable sets of individual classes. 
As mentioned in \cite{fan2008}, the selected variable sets for individual classes are not necessarily the same, thus we study the method that can identify  variable sets for individual classes.
We propose a variable detection process that can separately identify variable sets for estimating the density of individual classes,
and then integrate such a novel variable detection procedure into a nonparametric posterior kernel density classifier approach based on a density estimation method \citep{Sugiyama10a, Sugiyama12}.  {\color{black} We call the proposed classifier NPKDC-vd, which is the short for ``nonparametric posterior kernel density classifier with variable detection.'' }

In the rest of this paper, we discuss the theoretical properties of the proposed method, and show that this method can asymptotically reach the Bayes rule. 
Moreover, we describe its related algorithms including construction of classification rules, variable identification, and training sample size determination.
We explain how a general probability-based-model works in Section 2.1. 
We also explain how we can just use the relevant variables for this task, and how to determine the number of relevant variables in Section 2.2.  
In Section 2.3, we introduce the hypotheses testing ignorer to take the estimation variance into our procedure. 
Then, in Section 2.4, we explain how we can decide the training sample sizes of each class.  
The asymptotic property of the proposed method is in Section 2.5, and its proof is in the Supplementary.
We then illustrate our method with numerical studies with the synthesized and real data in Section 3 and 4, respectively, and then summarize our findings in the Summary section.

\section{Nonparametric Posterior Kernel Density Classifier with Variable Detection}
\label{sec:2}

\subsection{Nonparametric classifier}
We now describe the NPKDC-vd below.
Let $\mathbf{X}\subset \mathbb{R}^d$ be a $d$-dimensional vector of variables, and $Y \in  \{1, 2, \dots , c\}$ be the class label, where $c$ is the number of classes. Suppose that we do not have any prior prevalence information of each class; i.e. assume that $P(y=Y)=1/c$,  for $Y \in  \{1, 2, \dots , c\}$.
Assume that $\mathbf{X} \times Y$ have a joint probability density function $f_{\mathbf{X}, Y}(\mathbf{x}, y)$, and $\mathbf{x} \in \mathbf{X}$ has a marginal density $f_{\mathbf{X}}(\mathbf{x})$, then the class-posterior probability is  
$P(Y=y|\mathbf{X}=\mathbf{x}) ={f_{\mathbf{X}, Y}(\mathbf{x}, y)}/{f_{\mathbf{X}}(\mathbf{x})}$. 
It implies that a sample $\mathbf{x}$ belong to Class ${y}$ with probability $P(Y=y|\mathbf{X}=\mathbf{x})$,
and a probability-based classification rule will assign it to Class $\hat{y}$, if $\hat{y} := \argmax_{y \in \{1, \dots, c\}} P(Y=y|\mathbf{X}=\mathbf{x})$.  
To apply this idea we need to estimate the class-posterior probabilities,  
which is a challenging problem from both computational and theoretical perspectives; especially when the number of class $c$ and/or the dimension $d$ is large.  

Suppose that $\{\mathbf{x}_i, y_i)$, $i=1, \dots, n\}$, are i.i.d paired samples of size  $n$. 
Using the idea of {\color{black} the least-squares probabilistic classifier (LSPC)}  with a nonparametric kernel density estimation of conditional probabilities \citep[see][]{Sugiyama10a, Sugiyama12}, 
we have the class-posterior probability of Class $y$ as a summation of  kernel bases $K'$:
\begin{equation}
	P(Y=y|\mathbf{X}=\mathbf{x}; \bm{\alpha}) = \sum_{y'=1}^c\sum_{i=1}^n\alpha_i^{y'}K'(\mathbf{x}, \mathbf{x}_i, y, y'),
\label{LSPC0}
\end{equation}
which has $c \times n$ parameters: $\bm{\alpha}  = (\alpha_1^1, \dots, \alpha_n^1, \dots, \alpha_1^c, \dots, \alpha_n^c)^T$. 
Let  $\{\mathbf{x}_i^y, i = 1, \dots, n_y\}$ denote the samples in Class $y$ of size $n_y$.
Then via the delta kernel for $Y$ to separate $\mathbf{X}$ and $Y$, and a symmetric Gaussian kernel $\mathcal{K}$ on samples within the target class to reduce the number of kernels, we have the class-posterior probability for Class $y$ as follows:
\begin{equation}
	P(Y=y|\mathbf{X}=\mathbf{x}; \bm{\beta^y}) := \sum_{i=1}^{n_y}\beta_i^y \mathcal{K}(\mathbf{x}, \mathbf{x}_i^y),
\label{LSPC1}
\end{equation}
where $\beta^y$, $y=1, \ldots, c$, is an unknown vector to be estimated. 
Assume $\beta^y$, $y=1, \ldots, c$, are known, then assign sample $\mathbf{x}$ to Class $\hat{y}$, if
\begin{align*}
\hat{y} &:= \argmax_{y \in \{1, \dots, c\}} P(Y=y|\mathbf{X}=\mathbf{x}; \bm{\beta^y})
	 :=  \argmax_{y \in \{1, \dots, c\}} \sum_{i=1}^{n_y}\beta_i^y \mathcal{K}(\mathbf{x}, \mathbf{x}_i^y).
\end{align*}
Note that \eqref{LSPC1} suggests that we can decompose the original problem into $c$ independent problems, and use only the training samples of each class to estimate its own posterior probability. Because we do not use the whole training samples for an individual class, the number of parameters is equal to $\sum_{y=1}^c n_y$ parameters, which is smaller than that in \eqref{LSPC0}. Thus, using this decomposition can cut down the computational cost.
Moreover, via such decomposition, we can carry out this learning process with a parallel computing facility, or a machine with multi-core/multi-thread CPUs; that is, we can learn $c$ models simultaneously and separately.

\subsection{Bandwidth determination}
The conventional LSPC uses a prefixed bandwidth for its Gaussian kernels for each class.
This setting is not suitable when the density of each class has its own key variables, which
 is common in multiple-class classification problems \cite[see also][]{fan2008}.

We propose a bandwidth selection process and integrate it into \eqref{LSPC1} 
such that the estimates via \eqref{LSPC1} can have different smoothing parameters for individual coordinates of its kernel density estimation.
Then the class-posterior density becomes
\begin{align}
	P(Y=y|\mathbf{X}=\mathbf{x}; H_y) &:= \frac{1}{n_y} \sum_{i=1}^{n_y}\frac{1}{\det(H_y)}\mathcal{K}(H_y^{-1}(\mathbf{x} - \mathbf{x}_i^y))\nonumber\\
	& := \frac{1}{n_y}\sum_{i=1}^{n_y} \prod_{j=1}^d\frac{1}{h_j^y}K\Big(\frac{x_j - x_{ij}^y}{h_j^y}\Big),
\label{NPKDC-vd}
\end{align}
where $H_y = diag(h_1^y, \dots, h_d^y)$ is a diagonal matrix and its element $h_j^y$ is the bandwidth of one-dimensional Gaussian kernel $K$ in the $j$th coordinate for Class $y$. 
From \eqref{NPKDC-vd}, if  ${h_j^y}$ is small, then its correspondent variable will have larger contributions to the posterior density. 
For a probability-based classification rule, the samples in Class $y$ should have higher class-posterior probability for Class $y$. 
It follows that variables with smaller bandwidth values (i.e. $h_j^y$'s in \eqref{NPKDC-vd}) are more important to the posterior density of  Class $y$.
Then the classification rule of NPKDC-vd via \eqref{NPKDC-vd} becomes  
\begin{align}
\hat{y} 	=  \argmax_{y \in \{1, \dots, c\}} \frac{1}{n_y}\sum_{i=1}^{n_y} \prod_{j=1}^d\frac{1}{h_j^y}K\Big(\frac{x_j - x_{ij}^y}{h_j^y}\Big)\label{eq:BayesianNPKDC-vd}.
\end{align}
We then treat the bandwidths as some unknown parameters to be estimated with data, and will discuss its estimation procedure below.


\subsubsection{Greedy bandwidth estimation}
\cite{scott-sain}  claim that the direct estimation of the full density by kernel methods is only feasible when the dimension of variable vector is no larger than 6.
There are many criteria reported in the literature for determining bandwidth. However, as far as we know, most of them consider all variables equally and ignore the effects of the variables with no or little impact \citep{LEIVAMURILLO2012, Gu2013}, such that their methods suffer from the curse of dimensionality due to  too many redundant variables included in the estimating processes \cite[see also][]{fan2008}.  
The computational cost is one of the major issues for deciding the bandwidth for each dimension of data sets with lengthy variables. 
Thus, identifying high impact variable sets efficiently and effectively for density estimation of each class will not only largely improve the interpretation ability and classification performance, but reduce the computational cost.
Here we combine the idea of LSPC, and the regularization of derivative expectation operator (rodeo) \citep[see][]{pmlr-v2-liu07a} together in the proposed bandwidth selection  procedure such that we can detect the relevant variables with faster convergence rate and lower computational cost.

Without loss of generality and for simplicity of discussion, we rearrange the order of variables in each Class $y$, $y=1, \ldots, c$, such that the first $ r_y$ variables, $X_j: j =1 \leqslant j \leqslant r_y$, are relevant high impact variables for the given class, and $X_j: j=r_y+1 \leqslant j \leqslant d$ are the rest irrelevant ones.
(Please note that this arrangement is only for the convenience of discussion, and is not necessary in practice.)
Let $R_y = \{j: 1 \leqslant j \leqslant r_y \}$ and $R_y^c=\{j:r_y+1 \leqslant j \leqslant d\}$ be two index sets, where $r_y$ is the number of the relevant variables for Class $y$, $y=1, \ldots, c$, and this number can be different among classes. 
The details about determining the largest $r_y$ for the index set $R_y$ are in Section \ref{sec:feature-identification}, 
and Algorithm \ref{alg2} is for this purpose.

We may omit the subscript $y$ of $R_y$ and $r_y$ from our formulas in the following discussion when there is no ambiguity.
Assume, for the moment, that $r_y=r$ is the largest number of key variables for class density estimation, and the rest variables have little or no impact, 
then the posterior density function of Class $y$ can be rewritten as
\begin{align}
	P\big(Y=y|\mathbf{X} = \mathbf{x}; H_y\big) & = g_{Y|\mathbf{X}_R}\big(\mathbf{x}_R; H_y^R\big)u(\mathbf{x}_{R^c})
	= g_{Y|\mathbf{X}_R}(\mathbf{x}_R, H_y^R)\nonumber,
\end{align} 
where $u(\cdot)$ is a uniform function, $g_{Y|\mathbf{X}_R}$ depends only on the set $\mathbf{X}_R$, 
and $H_y^R = diag(h_1^y, \dots, h_r^y)$ is an $r\times r$ submatrix of $H_y$. 
Then \eqref{NPKDC-vd} becomes
\begin{align}
&P\big(Y=y|\mathbf{X} = \mathbf{x}; H_y\big) 
=  \frac{1}{n_y}\sum_{i=1}^{n_y} \prod_{j=1}^d\frac{1}{h_j^y}K\Big(\frac{x_j - x_{ij}^y}{h_j^y}\Big) \nonumber \\
&=\frac{1}{n_y}\sum_{i=1}^{n_y} \left[\prod_{j=1}^r\frac{1}{h_j^y}K\Big(\frac{x_j - x_{ij}^y}{h_j^y}\Big)\right] \left[\prod_{j=r+1}^d\frac{1}{h_j^y}K\Big(\frac{x_j - x_{ij}^y}{h_j^y}\Big)\right]  \label{uniform}\\
& \approx \frac{1}{n_y}\sum_{i=1}^{n_y} \left[\prod_{j=1}^r\frac{1}{h_j^y}K\Big(\frac{x_j - x_{ij}^y}{h_j^y}\Big)\right]. \label{approx}
\end{align}
The second term of \eqref{uniform} follows a uniform distribution by assumption.
Hence, we can use a large bandwidth value on $h_j, j = r+1, \dots, d$, as in the greedily bandwidth selection approach, 
to get a smooth kernel density function to estimate such a uniform function.  
Thus, we can use \eqref{approx} to approximate \eqref{uniform}.

Let  $\mathbf{x}=(x_i, \dots, x_d)^T$ be a sample of Class $y$.
Using \eqref{approx}, the estimated posterior density of $\mathbf{x}$ is
\begin{equation*}
	P\big(Y=y|\mathbf{X} = \mathbf{x}; H_y\big) =  \frac{1}{n_y}\sum_{i=1}^{n_y} \prod_{j=1}^d\frac{1}{h_j^y}K\Big(\frac{x_j - x_{ij}^y}{h_j^y}\Big).
\end{equation*}
To determine bandwidths, we start with a bandwidth matrix $H_y = \textit{diag}(h_0, \dots, h_0)$ with a large $h_0$ via the rodeo method,  
and compute derivatives, for $ 1 \leqslant j \leqslant d$,
\begin{align}
Z_j &= \frac{\partial P\big(Y=y|\mathbf{X} = \mathbf{x}; H_y\big)}{\partial h_j^y}
	= \frac{1}{n_y}\sum_{i=1}^{n_y} \frac{\partial}{\partial h_j^y}\left[\prod_{k=1}^d\frac{1}{h_k^y}K\Big(\frac{x_k - x_{ik}}{h_k^y}\Big)\right]\nonumber\\ 
	&\equiv \frac{1}{n_y}\sum_{i=1}^{n_y} Z_{ji}.
\end{align}
If $K$ is the Gaussian kernel, then $Z_j$ becomes
\begin{align}
Z_j &= \frac{1}{n_y}\sum_{i=1}^{n_y} Z_{ji}
	= \frac{1}{n_y}\sum_{i=1}^{n_y}\frac{(x_j - x_{ij})^2-(h_j^y)^2}{(h_j^y)^3} \prod_{k=1}^d\frac{1}{h_k^y}K\Big(\frac{x_k - x_{ik}}{h_k^y}\Big).
\end{align}
If the derivative $|Z_j|$ is small and the corresponding value of  $h_j^y$ is relatively large, it suggests that the variable $X_j$ may be irrelevant in the current density estimation. 
If $|Z_j|$ is large and to change $h_j^y$ will lead to a substantial change in its corresponding estimate, then we set the bandwidth equal to a smaller number   
{$\gamma \times h_j^y$}, where  {$\gamma \in (0, 1)$} is a small constant. 
We repeat this procedure for each $j$, $j=1, \ldots, d$, and keep shrinking its corresponding bandwidth in discrete steps, such as {$1, \gamma, \gamma^2, \dots$}, until the value of $|Z_j|$ is less than a given threshold $\lambda_j$.

To take the estimation variance into consideration, we use the sample variance of $Z_{j}$s, say $s_j^2$, to estimate $\sigma_j^2 \equiv \mbox{Var}(Z_j)$.
Following the suggestion of \cite{pmlr-v2-liu07a} and
to balance variance against bias,  we set a threshold $\lambda_j = s_j\sqrt{2\log(n_y c)}$ with $c = O(\log n_y)$. 
Once we have the posterior densities for all classes,  we use \eqref{eq:BayesianNPKDC-vd} as a rule to assign a class label for each sample.
Algorithm~\ref{alg1} describes how to estimate the posterior density estimate with bandwidth selection for each class. 
(Note that users can use other kernel functions with these algorithms via some modifications.)

\subsection{Variable identification and training sample size determination}
\label{sec:feature-identification}
We use the estimated bandwidths obtained from Algorithm~\ref{alg1} as indices of the relevant level of variables,
which depends on the estimate of posterior density of each class, and therefore  can be different among classes.
This novel variable determination method for constructing multiple-class classifiers can greatly improve 
the model-interpretation ability of classification rules, and provide more information applications or planning further studies.  

\subsection*{Identifying high impact variables}

We have the local bandwidths $\hat h_{j,i}^y$ of each variable $X_j$, $j = 1, \dots, d$, in each Class $y$ after applying Algorithm~\ref{alg1} to the training samples of each class.
As discussed before, if $X_j$ is a relevant variable, then it should have a smaller bandwidth compared with that of irrelevant variables in this class.
Because estimates of bandwidths depend on the samples, we take the variances of bandwidth estimates into consideration via adopting statistical hypotheses testing procedures to decide whether there are significant differences among the bandwidths of variables. We state this procedure below:

\noindent{\it Step 1:}
We first want to know whether all bandwidth means are equal, so the null and alternative hypotheses are  
\begin{align}
\label{anova}
	H_0 : ~& \bar h_1^y = \bar h_2^y = \dots = \bar h_d^y\\
	H_1 : ~& \text{not all bandwidth means are equal},\nonumber
\end{align}
where $\bar h_j^y = \sum_{i = 1}^{n_y}\hat h_{j,i}^y/n_y$, $ j = 1, \dots, d$. 
In our algorithm, all bandwidths start from the same initial value $h_0$.  
We use a one-way analysis of variance (ANOVA), and reject the null hypothesis if
\begin{equation}
\label{Ftest}
	F \equiv MS_B/MS_w > F_{\alpha, d-1, n_y\cdot d-d},
\end{equation}
where $MS_B$ is between-class mean square, $MS_w$ is within-class mean square, and $F_{\alpha, d-1, n_y \cdot d-d}$ is the upper $100 \cdot (1-\alpha)$th percentile of the F distribution with parameters $d-1$ and $n_y \cdot d-d$ degree of freedom.

\noindent{\it Step 2.}
Rejecting the null hypothesis in \eqref{anova} indicates that all bandwidth means are not the same. We then want to make sure which means are different via a post-hoc analysis through a multiple pairwise comparison:
\begin{align}
\label{Tukey}
	H_0 : ~& \bar h_m^y = \bar h_n^y, ~~ m,n = 1, ~\dots, d, m \neq n\\
	H_1 : ~& \bar h_m^y \neq \bar h_n^y.\nonumber
\end{align}
Because multiple testings are performed together, we adopt Tukey's honestly significant difference (Tukey's HSD) procedure, instead of $t$-test, based on the studentized range distribution \citep{Hochberg:1987:MCP:39892,10.2307/3001913}. Thus, the null hypothesis in \eqref{Tukey} is rejected if 
\begin{equation}
\label{post}
|t_{mn}| = \frac{|\bar h_m^y -\bar h_n^y|}{\sqrt{MS_w(\frac{2}{n_y})}} > \frac{1}{\sqrt{2}}q_{\alpha,d,n_y \cdot d-d},
\end{equation}
where $q_{\alpha,d,n_y \cdot d-d}$ is the upper $100(1-\alpha)$th percentile of the studentized range distribution with parameters $d$ and $n_y \cdot d - d$ degree of freedom. 
To identify relevant variables, we need to know exactly which variables have the bandwidth means which are significantly smaller than others,
To accelerate the testing process, we arrange the bandwidth means in an increasing order. 
Denote the sorted bandwidth means as $\bar h_{1^*}^y \leq \bar h_{2^*}^y \leq \dots \leq \bar h_{d^*}^y$, then we aim to find 
\begin{equation}
\label{vs}
o = \argmax_{m = 1,\dots, d} \Big\{|t_{m^*n^*}| = \frac{|\bar h_{m^*}^y -\bar h_{n^*}^y|}{\sqrt{MSE(\frac{2}{n_y})}} > \frac{1}{\sqrt{2}}q_{\alpha,d,n_y \cdot d-d}, \forall ~n > m\Big\}.
\end{equation}
Because after sorting, bandwidths $\bar h_{m^*}^y, ~m < o$ is smaller than $\bar h_{o^*}$, they are all significantly smaller than $\bar h_{n^*}, \forall n > o$.
It implies that the corresponding $\bar h_{o^*}$ is significantly smaller than $\bar h_{n^*}, \forall n > o$.
Thus, we choose the corresponding variables $X_{1^*}, \dots, X_{o^*}$ as the relevant variables in Class $y$ as follows:
\begin{equation}
j \in 
\begin{cases}
    R_y,& \text{if } \ \bar h_j^y \leqslant \bar h_{o^*}^y\\
    R_y^c,              & \text{otherwise}.
\end{cases}
\label{Eq:xr}
\end{equation}
We summarize this variable selection procedure as Algorithm~\ref{alg2}.
For convenience, we also put the classification and this class-specific variable determination together as Algorithm \ref{alg3}.  

The proposed approach emphasizes more on ``understanding how each sample is assigned to a class,'' or ``these samples belong to a particular class due to which variables.''
This is, of course, different from the conventional model interpretation of linear or other parametric models. 
However, this information is useful in some research/decision-making scenarios, such as medical/pharmaceutical studies, manufacturing, and so on.
Note that these algorithms are easily implemented with multi-core/multi-thread computers, or parallel computing facilities. Thus, they
will not dramatically increase the computational cost.

\begin{algorithm}[t]\fontsize{10}{8}\selectfont  
\SetAlgoLined
  \caption{Variable Selection for Individual Classes}
  \label{alg2}
    \KwData{
	\begin{itemize}
		\item	$\mathbf{x}_i=(x_{i1}, \dots, x_{id})^T, 
      i = 1, \dots, n_y$: training data set of Class $y$
	\end{itemize}
	}
    \KwIn{
	\begin{itemize}
		\item $\alpha$: significant level
	\end{itemize}
    }
    \KwOut {
	\begin{itemize}
	\item High impact index set $R_y$
	\end{itemize}
    }
        
    \emph{\textbf{Learning}}
    \begin{enumerate}
   	\item \For{$i = 1, \dots, n_y$}{
	Find the local bandwidths $\hat H_{y,i} = \text{diag}(\hat h_{1,i}^y, \dots, \hat h_{d,i}^y)$
	for data point $\mathbf{x}_i$ by Algorithm~\ref{alg1}
	}
	\item Calculate bandwidth means  $\bar h_j^y$, $j = 1, \dots, d$.
	\item one-way ANOVA test for equality of all bandwidth means by \eqref{anova} and \eqref{Ftest}
	\item Stop if null hypothesis \eqref{anova} is not rejected, otherwise continue
	\item Tukey's HSD for Post-Hoc Analysis by \eqref{Tukey} and \eqref{post}
	\item Find the largest mean $\bar h_{o*}^y$ that has significant different than all the larger ones by \eqref{vs}
	\item Decide $R^y$ by \eqref{Eq:xr}.
    \end{enumerate}
\end{algorithm}

\begin{algorithm}[t]\fontsize{10}{8}\selectfont  
\SetAlgoLined
  \caption{Integration of Classification and Variable Detection}
  \label{alg3}
    \KwData{
      $\{\mathbf{x}^{y'}\}=\{\mathbf{x}^{y'}_{1}, \dots, \mathbf{x}^{y'}_{n_{y'}}\}$, 
      ${y'}= 1, \dots, c$: training data set in $y'$th class; \\
      \hspace{42pt}$\{\mathbf{x}_i, y_i\}$ : $i = 1, \dots, m$ : testing data\\
	}
    \KwIn{$0 < \gamma < 1$: reduce rate for bandwidth, \\
    		\hspace{42pt}$h_0 = c_0/\log\log n$: initial bandwidth, \\
		\hspace{42pt}$\tau$: threshold}
    \KwOut {Estimated label: $\hat y_i$, $i = 1, \dots, m$,\\ 
    		\hspace{52pt}Selected variables: $R^{y'}$, $y' = 1, \dots, c$
    }

    \emph{\textbf{Learning}}\\    
    \For{$y'=1, \dots, c$}{
    	Use training data set $\{\mathbf{x}^{y'}\}$\\
    	\For{$i = 1, \dots, m$}{
		
		Estimate bandwidths $\hat H_{y',i}=(\hat h^{y'}_{i1}, \dots, \hat h^{y'}_{id})$ via Algorithm \ref{alg1} on testing data $\mathbf{x}_i$.\\
		Calculate the estimated density $\hat p(y=y'|\mathbf{x}_i;\hat H_{y',i})$
	}
    }
    
    \emph{Classification}\\
    	$\hat y_i = \argmax_{y'} \frac{\hat p(y=y'|\mathbf{x}_i;\hat H_{y',i})}{\sum_{y'=1}^c \hat p(y'|\mathbf{x}_i;\hat H_{y',i})}$, $i = 1, \dots, m$ \\

    Accuracy $= \{\sum_{i=1}^m I(\hat y_i = y_i)\}/m$, where $I(\cdot)$ is an indicator function.
    
    \emph{Variable Selection}\\
    \For{$y'=1, \dots, c$}{
    	($\hat h^{y'}_1, \dots, \hat h^{y'}_d) = \frac{\sum_{i=1}^mH_{y',i}\times I(\hat y_i = y')}{\sum_{i=1}^mI(\hat y_i = y')}$\\
    	Apply Variable Selection algorithm for Class $y'$ and get $R_{y'}$
    }

\end{algorithm}

\subsection{Training Sample Sizes of Individual classes}
In general, when training sample sizes among classes are highly imbalanced, we will usually obtain a biased classification rule with this training set.
On the other hand, a class with more relevant variables requires larger training sample size to estimate its density in our approach.
To use an equal sample size for all classes is a simple, but inefficient, approach. Thus, different classes require different training sample sizes, and 
to have appropriate training sample sizes among classes is important for multiple-class classification problems. 
To this end, we propose a two-step method for estimating the ``optimal'' sample sizes of individual classes based on Theorem \ref{thm1},  when a pilot study is allowed. 
We summarize the procedure as Algorithm \ref{alg4} in Supplementary,  for calculating training samples sizes.

Let $K$ be a $d$-dimensional bounded symmetric kernel (see also \eqref{eq:kernel}) satisfying  
\begin{equation}
	\int K(u)du = 1_d, \text{ and } \int uK(u)du = 0_d,
\end{equation}
and let $\nu$ and $\kappa$ be
\begin{equation}
	\nu = \int_{\mathbb{R}^d} uu^TK(u)du, \text{ and } \kappa = \Big(\int_{\mathbb{R}^d} K^2(u)du \Big)^{1/2}.
\end{equation}
Suppose that there is a  set of initial samples for all individual classes, then 
by Theorem \ref{thm1} (see  also \eqref{eq:errorrate3}), we have
\begin{align}
L_n  -  L^\star &\leqslant \sum_{y' = 1}^c \Bigg( \hat p_{y'}\int \Big| f_{y'}(\mathbf{x}) - \hat f_{y'}(\mathbf{x})\Big|dx + \int f_{y'}(\mathbf{x})\Big|p_{y'}  - \hat p_{y'} \Big|dx \Bigg)\nonumber\\
&=\sum_{y' = 1}^c \Bigg(\mathcal{O}\Big(n_{y'} ^{-2/(4+r_{y'})}\Big)+ \mathcal{O}\Big(\sqrt{\log(\log(n))/n}\Big)\Bigg). \nonumber
\end{align}   
It follows that the convergence rate of $ L_n - L^\star$ is asymptotically dominated by 
\begin{equation}\label{eq:p}
   \sum_{y' = 1}^c \hat p_{y'}\int \Big| f_{y'}(\mathbf{x}) - \hat f_{y'}(\mathbf{x})\Big|dx.
\end{equation}
Thus, for a given $\epsilon >0$, we have  
\begin{equation}
	\sum_{y' = 1}^c \frac{n_{y'}}{n}\Bigg(\frac{\nu}{2} \int \Big|\sum_{j=1}^{r_{y'}}k_j^{y'2}f_{y'}^{(jj)}\Big|  + \kappa \Big(\prod_{j=1}^{r_{y'}}k_j^{y'}\Big)^{-1/2}\int 	\sqrt{f_{y'}}\Bigg)n_{y'}^{-\frac{2}{4+r_{y'}}} \leqslant \epsilon.
\label{eq:samplesize1}
\end{equation}
Equation \eqref{eq:samplesize1} implies that
\begin{equation}
	\sum_{y' = 1}^c n_{y'}^{\frac{2+r_{y'}}{4+r_{y'}}}\Bigg(\frac{\nu}{2} \int \Big|\sum_{j=1}^{r_{y'}}k_j^{y'2}f_{y'}^{(jj)}\Big|  + \kappa \Big(\prod_{j=1}^{r_{y'}}k_j^{y'}\Big)^{-1/2}\int \sqrt{f_{y'}}\Bigg) \leqslant n\epsilon.
\label{eq:samplesize2}
\end{equation}
Let $\vec{A} = (A_1, \dots, A_c)'$, and $\vec{B} = (B_1, \dots, B_c)'$  be the vectors with components
\begin{align}
	A_{y'} = & n_{y'}^{(2+r_{y'})/(4+r_{y'})}, \text{ and }
\label{eq:A1}\\
	B_{y'} = & \frac{\nu}{2} \int \Big|\sum_{j=1}^{r_{y'}}k_j^{y'2}f_{y'}^{(jj)}\Big|  + \kappa \Big(\prod_{j=1}^{r_{y'}}k_j^{y'}\Big)^{-1/2}\int \sqrt{f_{y'}}, \;\; y' = 1 \dots, c,
\label{eq:B1}
\end{align}
respectively.   Then, $\sum_{y' = 1}^c A_{y'} B_{y'}  = \vec{A}\cdot\vec{B}\leqslant n \epsilon$
suffices to guarantee that $\eqref{eq:samplesize2} \leqslant n\epsilon$. 
This inner product $\vec{A}\cdot\vec{B}$ reaches its maximum, when $\vec{A}$ and $\vec{B}$ are parallel.
Hence, for a given $\epsilon >0$, if individual sample sizes satisfying $A \propto B$ with the total sample size equal to $n$,
then the difference of error rates between the proposed procedure and Bayes rule is approximately less than $\epsilon$.
Since the true density functions are unknown,  we use the estimated density functions instead.
Then, via Monte-Carlo integral method, we have the components of vectors $\vec A$ and $\vec B$ as follows:
\begin{align}
	\hat A_{y'} = & n_{y'}^{(2+\hat r_{y'})/(4+\hat r_{y'})}
\label{eq:A2}\\
	\hat B_{y'} = & \frac{\nu}{2} \sum_{\{\mathbf{x}^{\hat y'}\}} \frac{\Big|\sum_{j=1}^{\hat r_{y'}}k_j^{y'2}\hat f_{y'}^{(jj)}(\mathbf{x})\Big|}{\hat f_{y'}(\mathbf{x})}  + \kappa \Big(\prod_{j=1}^{\hat r_{y'}}k_j^{y'}\Big)^{-1/2}\sum_{\{\mathbf{x}^{\hat y'}\}} \frac{\sqrt{\hat f_{y'}(\mathbf{x})}}{\hat f_{y'}(\mathbf{x})}.
\label{eq:B2}
\end{align}
Thus,  with a given error bound, a pre-decided total sample size, and the information obtained from the first stage, 
we can adopt a two-stage procedure to estimate the required sample size of each class.

Algorithm \ref{alg4} in Supplementary (see Section \ref{ss})
describes a two-stage approach based on the above discussion, which includes two steps: {\it estimation and resampling}.
From the estimation step (E-step),  we have the estimated bandwidths for each class, and the estimated labels of samples. 
Using the estimated density and predicted label, we then use the formula above to decide whether to include 
more training samples in the resampling step (R-step), for the given error bound $\epsilon$ and total sample size $n$. 
Because the precision of the estimated sample sizes depends on the initial samples of individual classes, a reasonable size of initial samples for each class is required.

\subsection{Asymptotic properties and Algorithm of classification rule}
Let $\hat f_{y'}(\mathbf{x};\hat H_{y'})$ be the estimated probability density function using  rodeo method with the estimated bandwidth matrix $\hat H_{y'}=diag(\hat h_1^{y'},\dots,\hat h_d^{y'})$. 
Then we have the following theorem, and the proof of it is in  Supplementary \ref{Proof of Theorem 1}. 
\begin{thm}\label{thm1}
Assume the sample size in Class $y'$ is $n_{y'}$, and the total sample size is equal to $n = \sum_{y'\in \mathcal{Y}}n_{y'}$. 
Let $\hat p_{y'} = n_{y'}/n$ be the sample proportion,  and $\hat f_{y'}(\mathbf{x}; \hat H_{y'})$ be  the estimates via rodeo methiod.  
Then the classification rule: 
\begin{equation}
    \hat y = \argmax_{y' \in \mathcal{Y}}\hat p_{y'}\hat f_{y'}(\mathbf{x};\hat H_{y'})
\label{AppBayes}
\end{equation}
approximates to the Bayes rule: 
\begin{equation}
    y = \argmax_{y' \in \mathcal{Y}}p_{y'}f_{y'}(\mathbf{x}).
\label{Bayes}
\end{equation}
Let $L^\star$ and $L_n$ denote the probability of error rates of \eqref{Bayes} (the Bayes rule), and \eqref{AppBayes}, respectively. 
Then $L_n- L^*\rightarrow 0$ with probability one, as $\min_{y'=1, \ldots, c}\{n_{y'}\}$ goes to $\infty$.
\end{thm}
Algorithm \ref{alg1} describes the steps for conducting the NPKDC-vd -- short for the {\it nonparametric posterior kernel density classifier with the proposed bandwidth selection procedure}.

\begin{algorithm}[t]\fontsize{10}{8}\selectfont  
\SetAlgoLined
  \caption{Nonparametric Posterior Kernel Density Classifier with Variable Detection}
  \label{alg1}
    \KwData{
	\begin{itemize}
		\item	$\mathbf{x}_i=(x_{i1}, \dots, x_{id})^T$,  
      $i = 1, \dots, n_y$: training data set of Class $y$
      \item $\mathbf{x}$: a point on which we want to find the posterior density estimator
	\end{itemize}

	}
    \KwIn{
	\begin{itemize}
	\item \textcolor{black}{$0 < \gamma < 1$}: reduce rate for bandwidth
	\item $h_0 = c_0/\log\log n_y$: initial bandwidth for some constant $c_0$
	\item $c_n = O(\log n_y)$
	\end{itemize}
    }
    \KwOut {
	\begin{itemize}
	\item Bandwidths $\hat H_y = \text{diag}(\hat h_1^y, \dots, \hat h_d^y)$
	\item Posterior density estimator: $\hat P\big(Y=y|\mathbf{X} = \mathbf{x}; \hat H_y\big)$
	\end{itemize}
    }

    \bf{Initialization}\\

    $h_j^y= h_0$, $j = 1, \dots, d$\\
    $\mathcal{A} = \{1, 2, \dots, d\}$\\
    \vspace{0.1in}
    \While{$\mathcal{A}$ is nonempty}{
    	\For{$j \in \mathcal{A}$}{
		Estimate the derivative $Z_j$ and sample variance $s_j^2$.\\
		Compute the threshold $\lambda_j = s_j\sqrt{2\log(n_y c_n)}$.\\
		{If $|Z_j| > \lambda_j$, set $h_{j}^y \leftarrow \gamma h_{j}^y$;} otherwise remove $j$ from  $\mathcal{A}$.
	}
	Calculate the estimated posterior probability with the estimated $\hat H_y :\hat p (Y=y|x,\hat H_y)$
    }

\end{algorithm}


\section{Numerical Results: Synthesized Data}
\label{sec:7}
We use both synthetic and real datasets for illustration, and evaluate the performance based on both prediction power and variable detection ability. 
We set {\color{black} $c_0=10$, $c_n = \log n_y$, and $\gamma = 0.9$ as the default parameters for our algorithms, and also used the default parameter of Gaussian kernel ion Matlab.}
The baselines for comparison are the results obtained from classifiers without variable detection. 
To have a thorough assessment, we also use different numbers of classes, allocations among classes, and different training sample sizes of each class in the numerical studies.
We also apply the method of {\color{black} support vector machine (SVM)} to the same data sets for comparison purposes. 
We train a SVM based multiple-class classifier via the multiple-class error-correcting output codes (ECOC) model \cite[][]{Bagheri2012, Berger99}, which is trained with a bunch of the SVM-based binary classifiers with standardized predictors and Gaussian kernel function (with parameter equal to 1), and assign the testing data to class with the highest class posterior probabilities.  To illustrate the variable selection power of the proposed method and for comparison purposes, we adopt the method of {\color{black} ``SVM with sequential variable selection'' (SVMfs),} where variables are sequentially added to train a classification rule until there is no improvement in prediction \cite[see][for further details]{Bagheri2012, Berger99}.
SVMfs will only determine a subset, and the selected subset is shared with all classes, while our method allows each class to have its own subset of variables.

\subsection{Ten-class Example}
We generate a 10-class data set with 30 variables in this synthesized data set, 
and among them, each class has its own 6 relevant variables, which are not all the same among classes.
{\color{black}The relevant variables in the $y$th class are: $\{y, y+1, \dots, y+5\}$,} which are generated with distributions
\begin{equation*}
	X_i^y \sim \mathcal{N}(0.5, (0.02*(i-y+1))^2),\, \text{for } i = y : y+5,\, y = 1 : 10,
\end{equation*}
and the rest irrelevant variables are generated from 
\begin{equation*}
	X_i^y \sim \text{Uniform}(0, 1), \,\text{for } i \in \{1:30\}\setminus \{y:y+5\},\, y = 1 : 10.
\end{equation*}
{\color{black} The relevant variables of Class 1 are Variables 1 to 6, Class 2 has Variable 2 to 7 as its relevant variables, and so on.
That is, there are only some overlapping relevant variables among nearby classes.}
The training and testing sample sizes for each class are 150 and 100, respectively. 
We repeat this study 1000 times and Table \ref{Table1} shows the results based on 1000 trials.
We found that when the data set has some irrelevant variables, the proposed NPKDC-vd method has better classification accuracy ($67.52\%$) compared with that  of the conventional LSPC ($21.43\%$) at a price of computational time. 
The average  computational time of NPKDC-vd is around 100 seconds, 
SVMfs use around 140 seconds, while  LSPC, and SVM just take a few seconds. 
The sequential variable selection of SVMfs is based on an iterative algorithm, and that is the reason why it takes so much time. 
SVMfs perform better than SVM due to its variable selection procedure.  
The proposed NPKDC-vd method  has the highest accuracy in this example, which is much better than that of SVMfs.
As mentioned before, these SVM based methods rely on a bunch of SVM based binary classifiers.  
Thus, it is difficult to ``explain'' their classification rules.

\begin{figure}
\begin{center}
	\includegraphics[width=0.9\textwidth,height=0.8\textheight]{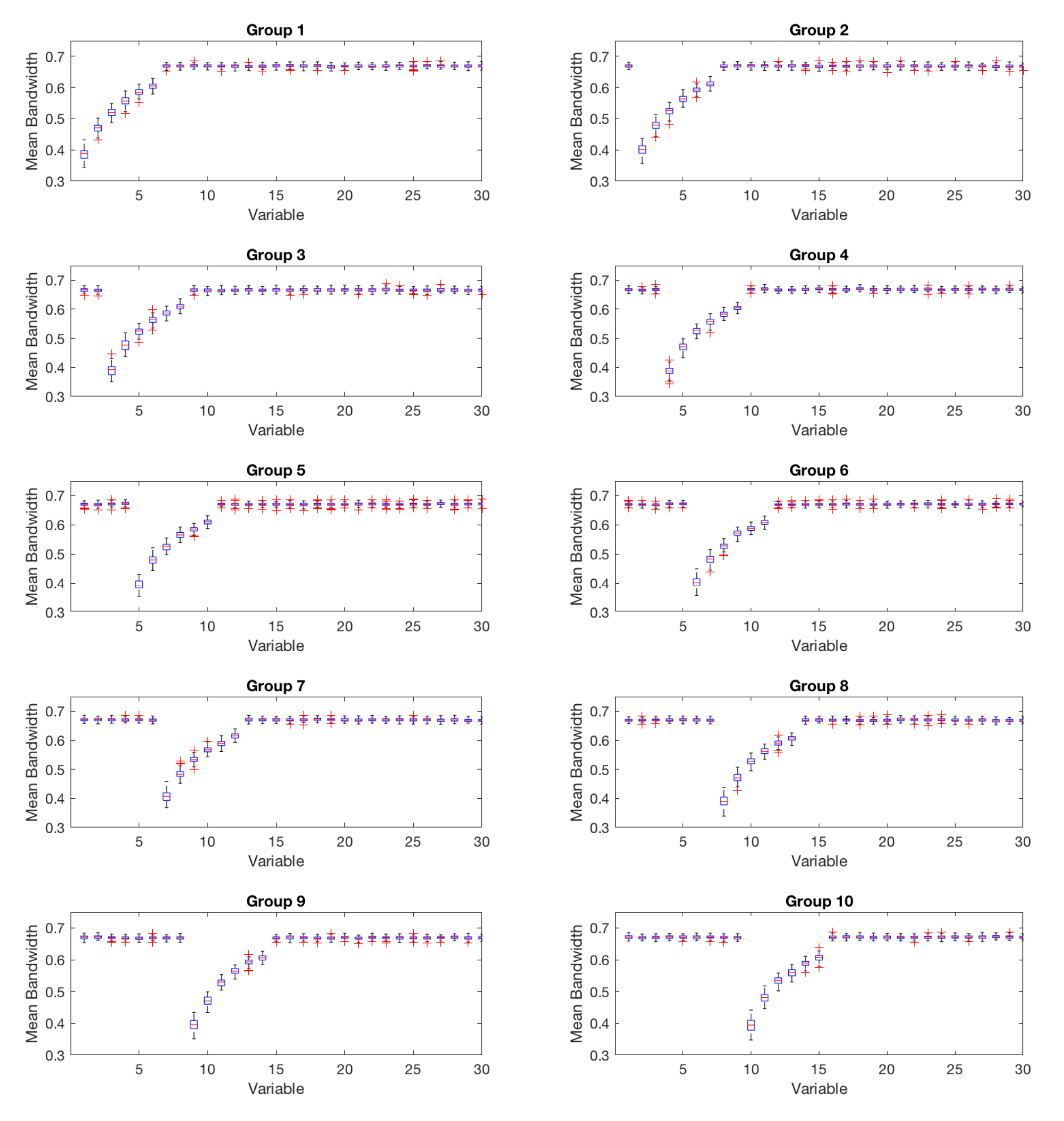}
	\caption{The box-plots of mean predicted bandwidths for the 10-class example. }
	\label{Figure1}
\end{center}
\end{figure}

Figure \ref{Figure1} shows the box-plots of the mean predicted bandwidths using the training samples in each class based on 1000 replications. 
In this plot, we can see that the mean bandwidths of the relevant variables are much smaller than those of the irrelevant ones, 
and the relevant variables for each class are separately identified. 
Table \ref{Table2} summaries the probabilities of relevant variables identified via hypotheses testing described in Algorithm \ref{alg2}.  
We highlight those relevant variables of each class with gray background color in this table, and
the probabilities in those gray cells are all equal to 1, and the rest are equal to 0.
Hence, the NPKDC-vd can identify the relevant variables precisely -- both false positive or false negative rates are 0.
It also confirms the predicted bandwidth is good guidance for identifying relevant variables. 
Table \ref{Table2_svm} is the variable identification probability of SVMfs.  We can see that the detection probability is much lower than that of NPKDC-vd.  Moreover, SVMfs only selects a set of variables for all classes, while NPKDC-vd can identify the relevant variables for each class with probability 1.  
Thus, the NPKDC-vd method provides more precise information about each class.
Moreover, we can see from this table that the computation of the posterior probability estimation with rodeo method via decomposition mentioned before, 
the increase in computational time is feasible (see NPKDC-vd and SVMfs in Table \ref{Table1}).
In fact, it can be shortened further with a parallel/distributed computation framework, since those estimates can be conducted separately.

\begin{table}
\begin{center}
\caption{ Classification results and computation time used in Example 1}
\label{Table1}
\begin{tabular}{ccc}\hline
 & \multicolumn{2}{c}{Method} \\ \cline{2-3} 
 & NPKDC-vd & LSPC \\ \hline 
Accuracy &   0.6752(  0.0167) &   0.2143(  0.0138) \\ \hline 
Time/sec & 101.8789(  1.5503) &   2.3559(  0.0540) \\ \hline 
\end{tabular}

\label{Table1_svm}
\begin{tabular}{ccc}\hline
  & \multicolumn{2}{c}{Method} \\ \cline{2-3} 
  & SVM & SVM(with fs) \\ \hline 
Accuracy &   0.1361(  0.0141) &   0.4655(  0.0690) \\ \hline
Time/sec & 3.7740(  0.1224) &   138.4383(  26.0866) \\ \hline
\end{tabular}
\end{center}
\end{table}

\begin{table}\fontsize{10}{8}\selectfont   
\caption{ The probabilities of relevant variables identified via NPKDC-vd for Example 1 based on 1000 iterations.}
\label{Table2}
 \begin{center}
\scalebox{0.65}{
\begin{tabular}{c|cccccccccc}\hline 
\diagbox{group~}{variable~~}& 1 & 2 & 3 & 4 & 5 & 6 & 7 & 8  & 9 & 10  \\ \hline \hline 
 1$\sim$10 & 0.1100\cellcolor[gray]{0.9} & 0.2000\cellcolor[gray]{0.9} & 0.2200\cellcolor[gray]{0.9} & 0.2900\cellcolor[gray]{0.9} & 0.3900\cellcolor[gray]{0.9} & 0.5000\cellcolor[gray]{0.9} &     0.4900\cellcolor[gray]{0.9} & 0.4100\cellcolor[gray]{0.9}  & 0.4400\cellcolor[gray]{0.9} & 0.3400\cellcolor[gray]{0.9}  \\ 

\hline \hline\diagbox{group~}{variable~~} & 11 & 12 & 13 & 14 & 15 & 16 & 17 & 18  & 19 & 20  \\ \hline \hline
 1$\sim$10 & 0.2700\cellcolor[gray]{0.9} & 0.1300\cellcolor[gray]{0.9} & 0.0900\cellcolor[gray]{0.9} & 0.0300\cellcolor[gray]{0.9} & 0.0100\cellcolor[gray]{0.9} & 0.0000 & 0.0000 & 0.0000  & 0.0000 & 0.0000  \\ 

\hline \hline\diagbox{group~}{variable~~} & 21 & 22 & 23 & 24 & 25 & 26 & 27 & 28  & 29 & 30  \\ \hline \hline
 1$\sim$10 & 0.0000 & 0.0000 & 0.0000 & 0.0000 & 0.0000 & 0.0000 &     0.0000 & 0.0000  & 0.0000 & 0.0000  \\ 
\hline \hline
\end{tabular}

}
 \end{center}
\end{table}

\begin{table}
\caption{ The probabilities of relevant variables identified by SVM with sequential variable selection based on 1000 iterations, for data in Example 1.}
\label{Table2_svm}
 \begin{center}
\scalebox{0.65}{
\begin{tabular}{c|cccccccccc}\hline 
\diagbox{group~}{variable~~}& 1 & 2 & 3 & 4 & 5 & 6 & 7 & 8  & 9 & 10  \\ \hline \hline 
 1$\sim$10 & 0.1100\cellcolor[gray]{0.9} & 0.2000\cellcolor[gray]{0.9} & 0.2200\cellcolor[gray]{0.9} & 0.2900\cellcolor[gray]{0.9} & 0.3900\cellcolor[gray]{0.9} & 0.5000\cellcolor[gray]{0.9} &     0.4900\cellcolor[gray]{0.9} & 0.4100\cellcolor[gray]{0.9}  & 0.4400\cellcolor[gray]{0.9} & 0.3400\cellcolor[gray]{0.9}  \\ 

\hline \hline\diagbox{group~}{variable~~} & 11 & 12 & 13 & 14 & 15 & 16 & 17 & 18  & 19 & 20  \\ \hline \hline
 1$\sim$10 & 0.2700\cellcolor[gray]{0.9} & 0.1300\cellcolor[gray]{0.9} & 0.0900\cellcolor[gray]{0.9} & 0.0300\cellcolor[gray]{0.9} & 0.0100\cellcolor[gray]{0.9} & 0.0000 & 0.0000 & 0.0000  & 0.0000 & 0.0000  \\ 

\hline \hline\diagbox{group~}{variable~~} & 21 & 22 & 23 & 24 & 25 & 26 & 27 & 28  & 29 & 30  \\ \hline \hline
 1$\sim$10 & 0.0000 & 0.0000 & 0.0000 & 0.0000 & 0.0000 & 0.0000 &     0.0000 & 0.0000  & 0.0000 & 0.0000  \\ 
\hline \hline
\end{tabular}
}
 \end{center}
\end{table}   


\subsection{Asymmetric Centers of Classes}
\label{Ex2}

\begin{figure}[h]
\begin{center}
	\includegraphics[width=12cm,]{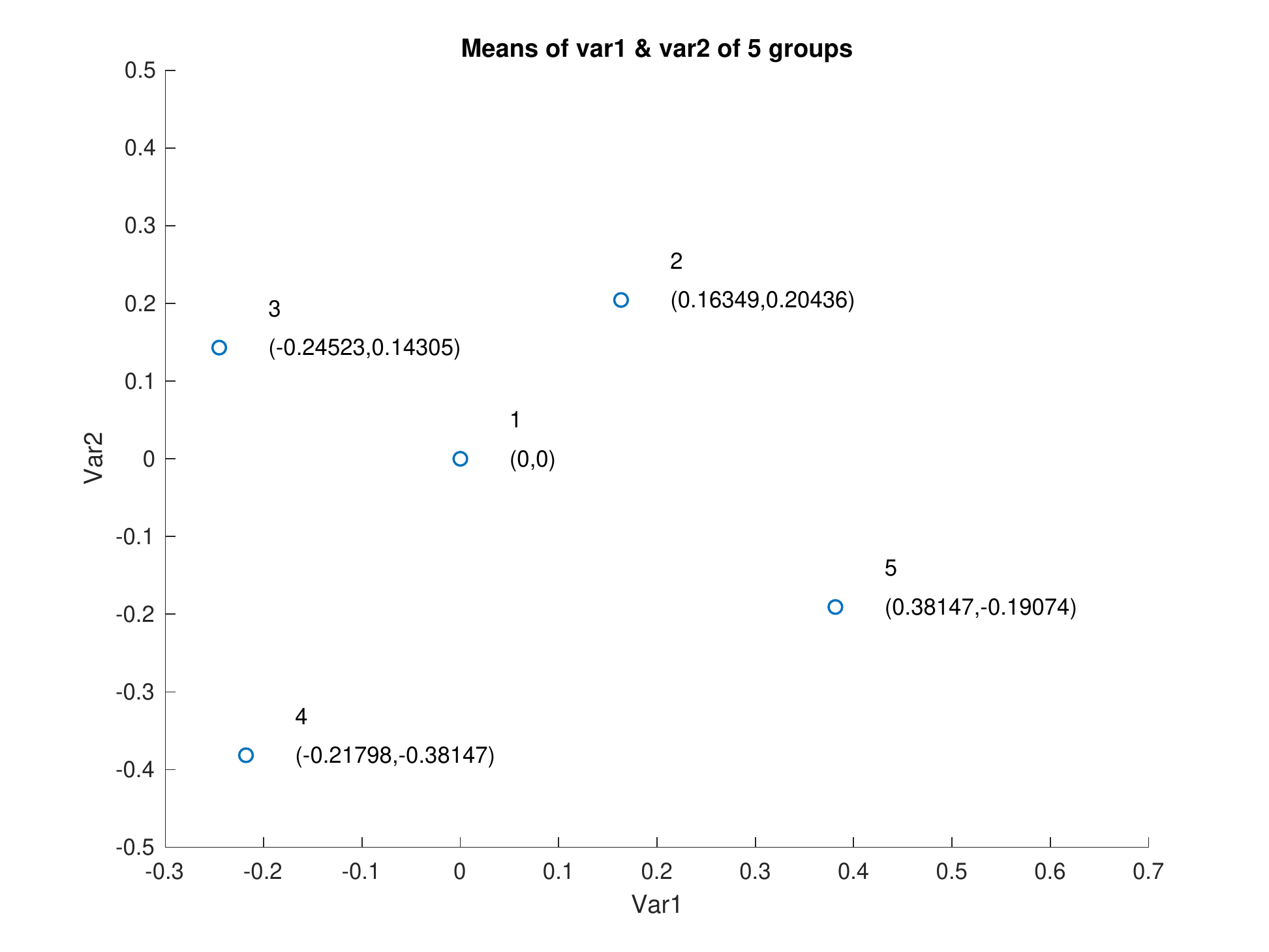}
	\caption{The means of 5 classes with asymmetric positions the first and second variables.}
	\label{Figure2}
\end{center}
\end{figure}

We now study the case with a complicated relative positions among classes.
To illustrate this situation, we generate a 5-class data set with 10 variables, and only the first two variables are relevant in all classes.
We generate 8  irrelevant variables from a uniform distribution, and the other two relevant variables from  \eqref{2-relevant} below:
\begin{equation}
	N\bigg(\begin{bmatrix}  \mu^y_1 \\ 
	\mu^y_2 \end{bmatrix}, \begin{bmatrix} 0.1^2 & 0 \\ 0 & 0.2^2 \end{bmatrix}\bigg), y = 1:5,
	\label{2-relevant}
\end{equation}
where $\mu^y_1$ and $\mu^y_2$ are location parameters.
Figure \ref{Figure2}  shows the values of the class means in the space spanned by these two relevant variables.
It shows that Class 1 (i.e. $y=1$) is surrounded by the other 4 classes in this space with different distances between their means.

\begin{figure} [t]
\begin{center}
	\includegraphics[width=0.7\textwidth]{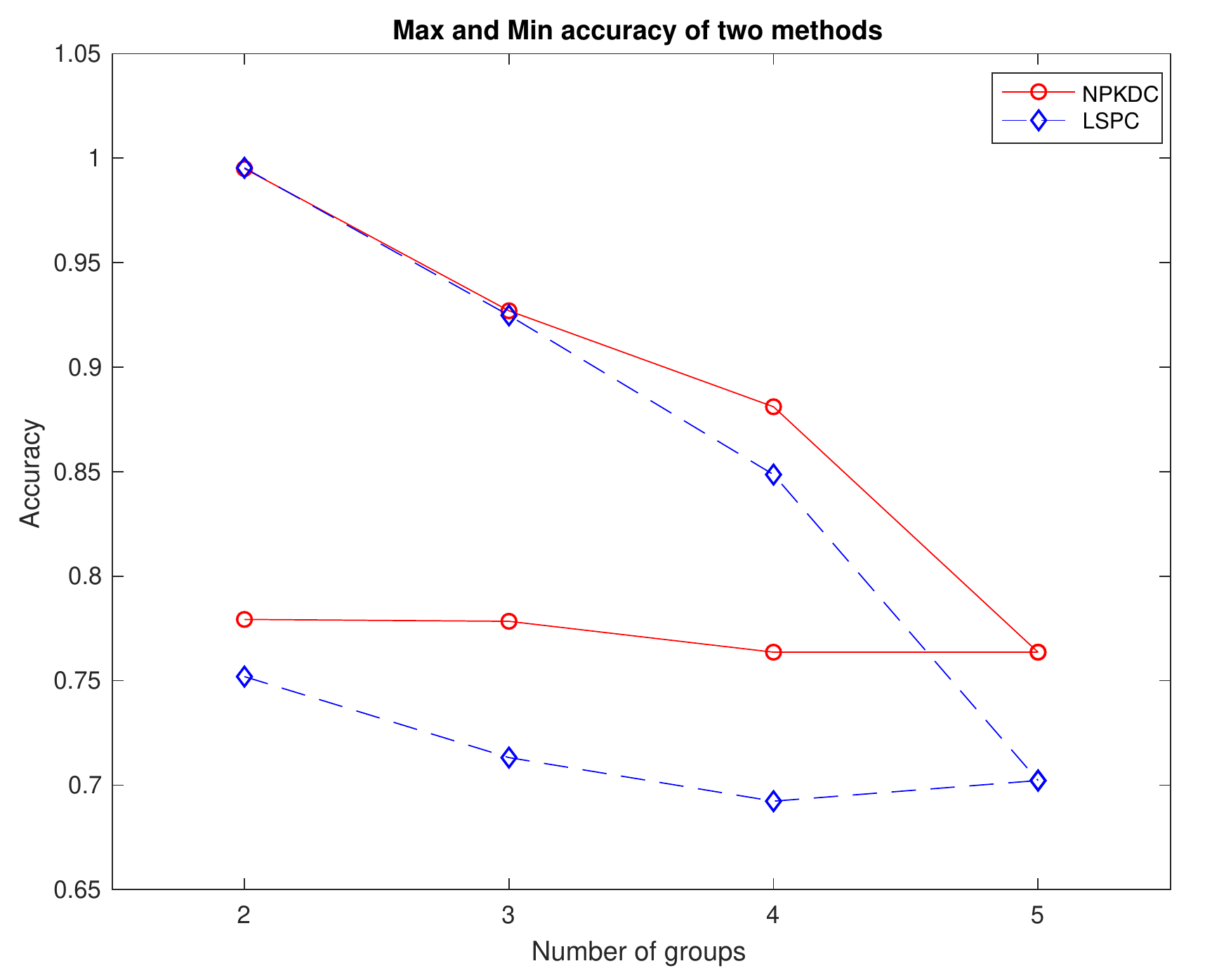}
	\caption{The maximum and minimum accuracies of two methods based on the data with asymmetric positions among classes.}
	\label{Figure3}
\end{center}
\end{figure}

We conduct studies with different combinations of these 5 classes to see how the number of classes, and their relative positions affect classification performance. 
{\color{black}The training and testing data sizes of each class are 150 and 100, respectively.
Table \ref{Table3} reports the classification accuracy based on 1000 replications of each study. }
We can see that the classification performance declines for all 4 methods when the number of classes increases,
since when the number of classes increases, the relations among classes are getting complicated such that the decision boundaries among classes are vague.   
From this table, we can also see that the performances are not all the same among the studies with the same number of classes, which is due to the relative positions among classes being all different.  
We can find a similar situation in Table \ref{Table3_svm}, 
which shows that the classification accuracy of SVM and SVMfs declines for all 4 methods when the number of classes increases.
Comparing these two tables, we found that SVMfs has the highest accuracy.  
NPKDC-vd has slightly lower accuracy than that of SVMfs
In general, methods with variable selection perform better than their own counterparts.

Table~\ref{Table4} shows the  variable detection probabilities,
and the probabilities of correctly selecting the first two relevant are equal to 1 for both NPKDC-vd and SVMfs.
(Note that the results for other cases are similar, and therefore are omitted.)
However,  the false positive probabilities of NPKDC-vd are less than $1\%$, 
while SVMfs has falsely selected probabilities of irrelevant variables larger than $10\%$ in all multiple-class setups in this example.
These results show that using estimated bandwidths for variable selection is promising.
It confirms that NPKDC-vd can retain both satisfactory classification accuracy, 
and reveal the information of relevant variables of individual classes, which improves the interpretation ability of the corresponding classification rules.

Figure \ref{Figure3} is a plot of the maximum and minimum accuracies of NPKDC-vd and LSPC, which shows that NPKDC-vd has higher accuracy than that of LSPC for all cases in this example.  
The maximum  accuracy declines  as the number of classes increases for both methods .
However, the minimum accuracy of NPKDC-vd retains at the level of accuracy (around 0.77) as the number of classes increases,
while that of LSPC drops from 0.75 to 0.70.


\begin{table}  [t]  
        \begin{center}
  \caption{ Classification accuracy on different combinations of classes in example with special located between classes in Section \ref{Ex2}. The training size = 150, and  testing size = 100 in each class. Results are based on 1000 repeated trials.}
        \label{Table3}
	\scalebox{0.63}{
       \begin{tabular}{c|c|c||c|c|c}\hline 
\# of classes & \multicolumn{2}{c||}{2} & \# of classes & \multicolumn{2}{c}{3} \\ \hline 
\diagbox{classes~}{Method~~} & NPKDC-vd& LSPC & \diagbox{classes~}{Method~~} & NPKDC-vd & LSPC \\ \hline 
        1        2 &   0.7793(  0.0287) &   0.7520(  0.0291) &        1        2        3 &   0.7784(  0.0232) &  0.7225(  0.0264) \\ \hline 
        1        3 &   0.8813(  0.0245) &   0.8315(  0.0300) &        1        2        4 &   0.7847(  0.0229) &  0.7132(  0.0259) \\ \hline 
        1        4 &   0.9004(  0.0211) &   0.8876(  0.0218) &        1        2        5 &   0.7898(  0.0246) &  0.7602(  0.0244) \\ \hline 
        1        5 &   0.9613(  0.0150) &   0.9431(  0.0185) &        1        3        4 &   0.8232(  0.0207) &  0.7993(  0.0229) \\ \hline 
        2        3 &   0.9676(  0.0118) &   0.9605(  0.0133) &        1        3        5 &   0.8945(  0.0174) &  0.7804(  0.0278) \\ \hline 
        2        4 &   0.9877(  0.0079) &   0.9720(  0.0131) &        1        4        5 &   0.9095(  0.0166) &  0.8672(  0.0177) \\ \hline 
        2        5 &   0.9023(  0.0244) &   0.8817(  0.0225) &        2        3        4 &   0.9074(  0.0163) &  0.9002(  0.0158) \\ \hline 
        3        4 &   0.8997(  0.0209) &   0.9070(  0.0209) &        2        3        5 &   0.9130(  0.0160) &  0.8570(  0.0194) \\ \hline 
        3        5 &   0.9950(  0.0046) &   0.9953(  0.0043) &        2        4        5 &   0.9270(  0.0138) &  0.9120(  0.0143) \\ \hline 
        4        5 &   0.9932(  0.0063) &   0.9929(  0.0063) &        3        4        5 &   0.9259(  0.0138) &  0.9248(  0.0145) \\ \hline 
\multicolumn{6}{c}{ } \\ \hline 
\# of classes & \multicolumn{2}{c||}{4} & \# of classes & \multicolumn{2}{c}{5} \\ \hline 
\diagbox{classes~}{Method~~} & NPKDC-vd & LSPC & \diagbox{classes~}{Method~~} & NPKDC-vd & LSPC \\ \hline 
        1        2        3        4 &   0.7636(  0.0223) &   0.7022(  0.0203) &        1        2        3        4        5 &   0.7728(  0.0211) &  0.6987(  0.0201) \\ \hline 
       1        2        3        5 &   0.7901(  0.0204) &   0.6924(  0.0226) & & &\\ \hline 
       1        2        4        5 &   0.7948(  0.0182) &   0.7314(  0.0202) & & &\\ \hline 
       1        3        4        5 &   0.8498(  0.0166) &   0.7636(  0.0177) & & &\\ \hline 
       2        3        4        5 &   0.8810(  0.0167) &   0.8486(  0.0192) & & &\\ \hline 
\end{tabular}

        }
        \end{center}
\end{table}

\begin{table}  [t]  
        \begin{center}
  \caption{ Classification accuracy for different combinations of classes in Section \ref{Ex2}. 
  Results are based on 1000 replications, and training and testing sizes are equal to 150, and 100, respectively, in each class. 
  SVM and SVMfs denote the SVM methods with/without variable selection preprocess.}
        \label{Table3_svm}
	\scalebox{0.63}{
       \begin{tabular}{c|c|c||c|c|c}\hline 
\# of groups & \multicolumn{2}{c||}{2} & \# of groups & \multicolumn{2}{c}{3} \\ \hline 
\diagbox{combinations~}{Method~~} & SVM & SVMfs& \diagbox{combinations~}{Method~~} & SVM & SVMfs \\ \hline 
        1        2 &   0.7981(  0.0301) &   0.8843(  0.0190) &        1        2        3 &   0.7725(  0.0264) &  0.8869(  0.0150) \\ \hline 
        1        3 &   0.8548(  0.0282) &   0.9470(  0.0149) &        1        2        4 &   0.8083(  0.0207) &  0.8954(  0.0159) \\ \hline 
        1        4 &   0.8962(  0.0223) &   0.9617(  0.0102) &        1        2        5 &   0.8036(  0.0226) &  0.8977(  0.0145) \\ \hline 
        1        5 &   0.9369(  0.0185) &   0.9900(  0.0051) &        1        3        4 &   0.8044(  0.0247) &  0.9178(  0.0156) \\ \hline 
        2        3 &   0.9347(  0.0220) &   0.9958(  0.0036) &        1        3        5 &   0.8683(  0.0168) &  0.9551(  0.0103) \\ \hline 
        2        4 &   0.9793(  0.0116) &   0.9974(  0.0032) &        1        4        5 &   0.8957(  0.0186) &  0.9666(  0.0095) \\ \hline 
        2        5 &   0.9052(  0.0254) &   0.9610(  0.0108) &        2        3        4 &   0.8822(  0.0219) &  0.9615(  0.0089) \\ \hline 
        3        4 &   0.8508(  0.0300) &   0.9486(  0.0141) &        2        3        5 &   0.9046(  0.0157) &  0.9702(  0.0093) \\ \hline 
        3        5 &   0.9735(  0.0156) &   0.9980(  0.0030) &        2        4        5 &   0.9167(  0.0156) &  0.9731(  0.0067) \\ \hline 
        4        5 &   0.9621(  0.0169) &   0.9982(  0.0030) &        3        4        5 &   0.9022(  0.0137) &  0.9651(  0.0082) \\ \hline 
\multicolumn{6}{c}{ } \\ \hline 
\# of groups & \multicolumn{2}{c||}{4} & \# of groups & \multicolumn{2}{c}{5} \\ \hline 
\diagbox{combinations~}{Method~~} & SVM & SVMfs & \diagbox{combinations~}{Method~~} & SVM & SVMfs \\ \hline 
        1        2        3        4 &   0.7606(  0.0203) &   0.8814(  0.0172) &        1        2        3        4        5 &   0.7806(  0.0163) &  0.8871(  0.0149) \\ \hline 
       1        2        3        5 &   0.7881(  0.0199) &   0.8944(  0.0153) & & &\\ \hline 
       1        2        4        5 &   0.8086(  0.0186) &   0.9034(  0.0133) & & &\\ \hline 
       1        3        4        5 &   0.8295(  0.0186) &   0.9325(  0.0109) & & &\\ \hline 
       2        3        4        5 &   0.8720(  0.0161) &   0.9496(  0.0096) & & &\\ \hline 
\end{tabular}
      
        }
        \end{center}
\end{table}

\begin{table}  [t]
	\caption{The probabilities of relevant variable identification, based on 1000 iterations, for the example with asymmetric positions among classes of NPKDC-vd and SVM with sequential variable selection.}
        \label{Table4_all}
        \begin{subtable}{0.9\linewidth}\tiny
	\centering
	\caption{Probabilities of the relevant variables identified via NPKDC-vd}
        \begin{tabular}{cc|>{\columncolor[gray]{0.9}}c>{\columncolor[gray]{0.9}}ccccccccc}\hline 
 \multicolumn{2}{c|}{\diagbox{\# of groups/\\group}{variables}} & 1 & 2 & 3 & 4 & 5 & 6 & 7 & 8 & 9 & 10 \\ 
 \hline 
 \multirow{2}{*}{2} & 1 &     1.00 &     1.00 &     0.01 &     0.00 &     0.00 &     0.00 &     0.00 &     0.00  &     0.00 &     0.00 \\
 & 2 &     1.00 &     1.00 &     0.00 &     0.00 &     0.00 &     0.00 &     0.00 &     0.00  &     0.00 &     0.00\\
 \hline 
 \multirow{3}{*}{3} & 1 &     1.00 &     1.00 &     0.00 &     0.00 &     0.00 &     0.00 &     0.00 &     0.00  &     0.00 &     0.00 \\ 
 & 2 &     1.00 &     1.00 &     0.00 &     0.00 &     0.00 &     0.00 &     0.00 &     0.00  &     0.00 &     0.01\\ & 3 &     1.00 &     1.00 &     0.00 &     0.00 &     0.00 &     0.00 &     0.00 &     0.00  &     0.00 &     0.00\\
 \hline 
 \multirow{4}{*}{4} & 1  &     1.00 &     1.00 &     0.00 &     0.00 &     0.00 &     0.00 &     0.00 &     0.00  &     0.00 &     0.00 \\ 
 & 2 &     1.00 &     1.00 &     0.00 &     0.00 &     0.00 &     0.00 &     0.00 &     0.00  &     0.00 &     0.00\\ 
 & 3 &     1.00 &     1.00 &     0.01 &     0.00 &     0.00 &     0.00 &     0.00 &     0.00  &     0.00 &     0.00\\ 
 & 4 &     1.00 &     1.00 &     0.00 &     0.00 &     0.00 &     0.00 &     0.00 &     0.00  &     0.00 &     0.00\\
 \hline 
 \multirow{5}{*}{5} & 1  &     1.00 &     1.00 &     0.00 &     0.00 &     0.01 &     0.00 &     0.00 &     0.00  &     0.00 &     0.00 \\ 
 & 2 &     1.00 &     1.00 &     0.00 &     0.00 &     0.01 &     0.00 &     0.00 &     0.00  &     0.00 &     0.00\\ 
 & 3 &     1.00 &     1.00 &     0.00 &     0.00 &     0.00 &     0.00 &     0.00 &     0.00  &     0.00 &     0.00\\ 
 & 4 &     1.00 &     1.00 &     0.00 &     0.00 &     0.01 &     0.00 &     0.00 &     0.00  &     0.00 &     0.00\\ 
 & 5 &     1.00 &     1.00 &     0.00 &     0.01 &     0.00 &     0.00 &     0.00 &     0.00  &     0.00 &     0.00\\
 \hline 
\end{tabular}

        \label{Table4}

	\caption{Probabilities of the relevant variables identified via SVM with variable selection preprocess}
        \label{Table4_svm}
        \begin{tabular}{cc|>{\columncolor[gray]{0.9}}c>{\columncolor[gray]{0.9}}ccccccccc}\hline 
 \multicolumn{2}{c|}{\diagbox{\# of groups/\\group}{variables}} & 1 & 2 & 3 & 4 & 5 & 6 & 7 & 8 & 9 & 10 \\ 
 \hline 
 2 & 1$\sim$ 2 &     1.00 &     1.00 &     0.16 &     0.17 &     0.14 &     0.16 &     0.21 &     0.20  &     0.21 &     0.19 \\
 \hline 
 3 & 1$\sim$ 3 &     1.00 &     1.00 &     0.23 &     0.21 &     0.23 &     0.22 &     0.20 &     0.25  &     0.18 &     0.09 \\
 \hline 
 4 & 1$\sim$ 4  &     1.00 &     1.00 &     0.10 &     0.15 &     0.23 &     0.17 &     0.20 &     0.18  &     0.22 &     0.15 \\
 \hline  5 & 1$\sim$ 5  &     1.00 &     1.00 &     0.21 &     0.16 &     0.27 &     0.14 &     0.18 &     0.20  &     0.22 &     0.12 \\
 \hline 
\end{tabular}

\end{subtable}
\end{table}    


\subsection{Training samples sizes and number of classes}
\begin{figure}[t]
	\begin{center}
	\includegraphics[width=0.7\textwidth]{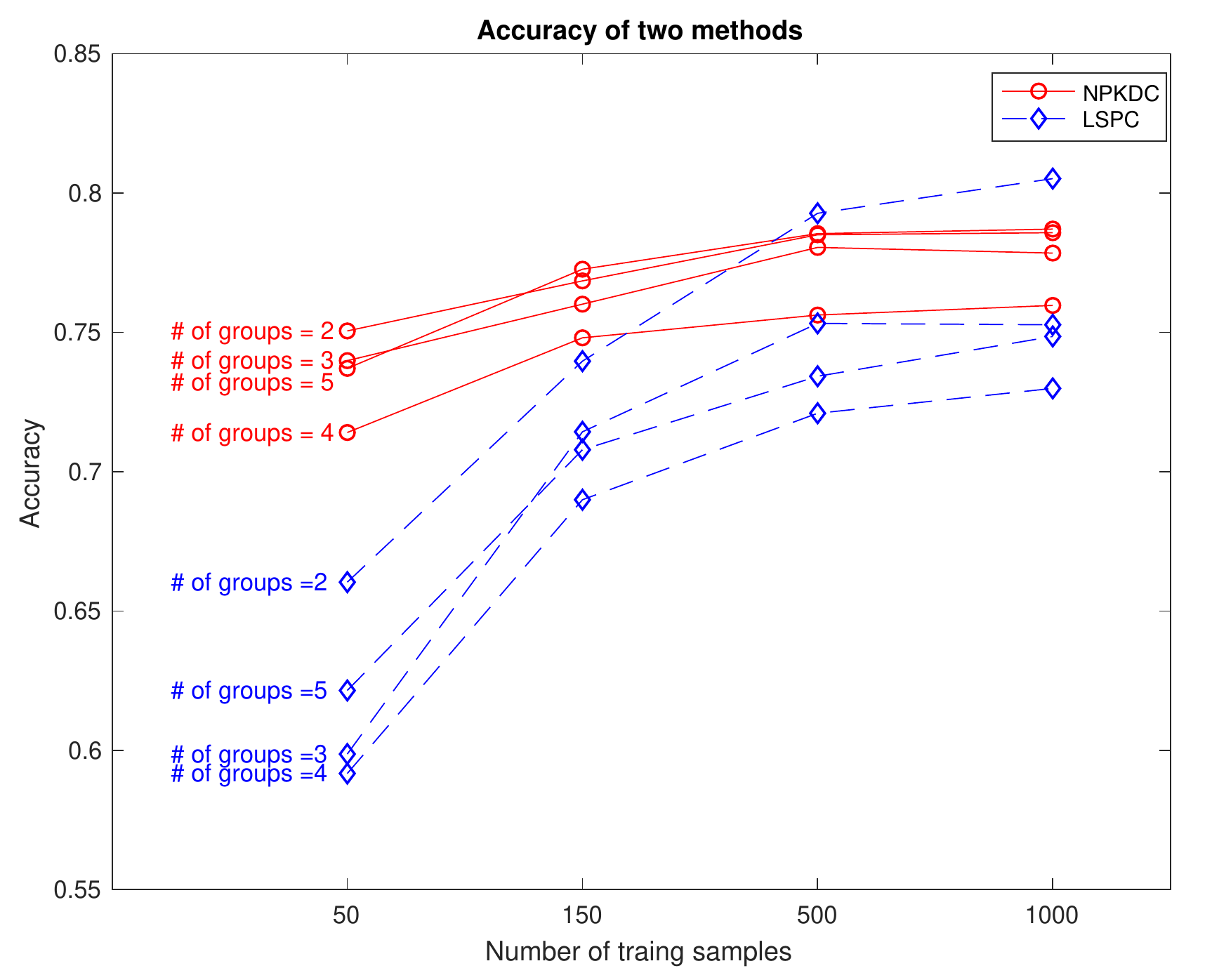}
	\caption{The accuracy rates of two methods.}
	\label{Figure4}
	\end{center}
\end{figure}

To illustrate how training data sizes affect the performance of NPKDC-vd and LSPC classifiers,
we generate  data sets  under the same setup as in Section \ref{Ex2}, but with different sizes of training sets,
where class sizes are 50, 150, 500, and 1000, and the testing sample size is 100 for each class.
We consider classification problems with 2 to 5 classes. 
Because the results are similar, when for the case with the same number of classes, 
we only report the first one for each setting as their representatives.
(For example, for two-class cases, we only report the results of Class 1 versus Class 2 , and omit the other two-class cases.)
Table \ref{Table6} summarizes the accuracy of this study, and Figure \ref{Figure4} shows the corresponding plots. 
The accuracy of both methods increases when the training size increases, and
the differences between the two methods become smaller when the training sample size gets larger.
When the training sample size is small,  NPKDC-vd clearly outperforms LSPC.
It is worth  noting that the NPKDC-vd outperforms LSPC in all cases with a number of classes larger than 2, 
and maintains a stable accuracy even when the training sample sizes are small.   
The LSPC can only be better in binary classification cases with more than 500 training samples. 
Besides these cases, the NPKDC-vd has about $10\%$ higher accuracy than that of the LSPC.

Table \ref{Table8} lists the probabilities of detecting the two relevant variables for all different sample-size cases. 
When the training sample size is no less than 150 for each class, the relevant variable detection rate of NPKDC-vd is 100\% in this example.
When there are only 50 training samples in each class, the detection rates for the second variable are less than 1 in these studies.  
Please note that in this table, we simply use variables 1 and 2 to denote the first and second relevant variables of each class, and they are not the same among classes.

\begin{table} [t]  
        \begin{center}
          \caption{Classification accuracies of NPKDC-vd and LSPC  based 1000 trails with different training examples sizes.}
        \label{Table6}
	\scalebox{0.6}{
       \begin{tabular}{c|cccc|cccc}\hline 
 & \multicolumn{8}{c}{Methods}  \\  
 & \multicolumn{4}{c}{NPKDC} & \multicolumn{4}{c}{LSPC} \\ \hline \hline
\diagbox{\# of training~}{\# of groups~~} & 2 & 3 & 4 & 5 & 2 & 3 & 4 & 5 \\ \hline
 \multirow{2}{*}{\bf 50}  & \cellcolor[gray]{0.9} 0.7505 & \cellcolor[gray]{0.9}  0.7398 & \cellcolor[gray]{0.9}  0.7141 & \cellcolor[gray]{0.9}  0.7371 &  \cellcolor[gray]{0.9} 0.6603 & \cellcolor[gray]{0.9}  0.5987 &  \cellcolor[gray]{0.9} 0.5917 & \cellcolor[gray]{0.9}  0.6215 \\ 
 & \cellcolor[gray]{0.9} (  0.0373)& \cellcolor[gray]{0.9}(  0.0282)& \cellcolor[gray]{0.9}(  0.0274)& \cellcolor[gray]{0.9} (  0.0240)& \cellcolor[gray]{0.9} (  0.0421)& \cellcolor[gray]{0.9} (  0.0354)& \cellcolor[gray]{0.9} (  0.0323)& \cellcolor[gray]{0.9}(  0.0304)\\\hline 
 \multirow{2}{*}{150}  &   0.7685 &   0.7601 &   0.7480 &   0.7726 &   0.7396 &   0.7143 &   0.6899 &   0.7079 \\ 
 & (  0.0305)& (  0.0255)& (  0.0212)& (  0.0198)& (  0.0325)& (  0.0245)& (  0.0203)& (  0.0184)\\\hline 
 \multirow{2}{*}{500}  &   0.7850 &   0.7805 &   0.7562 &   0.7854 &   0.7927 &   0.7532 &   0.7210 &   0.7342 \\ 
 & (  0.0296)& (  0.0220)& (  0.0191)& (  0.0171)& (  0.0286)& (  0.0230)& (  0.0186)& (  0.0164)\\\hline 
 \multirow{2}{*}{1000}  &   0.7857 &   0.7784 &   0.7596 &   0.7870 &   0.8052 &   0.7527 &   0.7299 &   0.7485 \\ 
 & (  0.0282)& (  0.0233)& (  0.0222)& (  0.0177)& (  0.0252)& (  0.0228)& (  0.0188)& (  0.0193)\\\hline 
\end{tabular}
        }
    \end{center}
\end{table}

\begin{table}[h]\footnotesize
        \begin{center}
        \caption{The  relevant variable identification probabilities of the NPKDC-vd with different training sample sizes and number of groups  based on 1000 iterations.}
        \label{Table8}

	\scalebox{0.8}{
        \begin{tabular}{cc||cc|cc|cc|cc}\hline 
 & & \multicolumn{8}{c}{\# of training samples}  \\  
 & & \multicolumn{2}{c}{50} & \multicolumn{2}{c}{150} & \multicolumn{2}{c}{500} & \multicolumn{2}{c}{1000} \\ 
 \hline
 \multicolumn{2}{c||}{\diagbox{\# of groups/\\group}{variables\\}} & 1 & 2 & 1 & 2 & 1 & 2 & 1 & 2 \\ 
 \hline 
 \multirow{2}{*}{2} & 1 &     1.00 &     0.85 &     1.00 &     1.00 &     1.00 &     1.00 &     1.00 &     1.00\\& 2 &     1.00 &     0.88 &     1.00 &     1.00 &     1.00 &     1.00 &     1.00 &     1.00\\
 \hline
 \multirow{3}{*}{3} & 1 &     1.00 &     0.88 &     1.00 &     1.00 &     1.00 &     1.00 &     1.00 &     1.00\\ & 2 &     1.00 &     0.94 &     1.00 &     1.00 &     1.00 &     1.00 &     1.00 &     1.00\\ & 3 &     1.00 &     0.77 &     1.00 &     1.00 &     1.00 &     1.00 &     1.00 &     1.00\\
 \hline
 \multirow{4}{*}{4} & 1  &     1.00 &     0.90 &     1.00 &     1.00 &     1.00 &     1.00 &     1.00 &     1.00\\ & 2 &     1.00 &     0.88 &     1.00 &     1.00 &     1.00 &     1.00 &     1.00 &     1.00\\ & 3 &     1.00 &     0.86 &     1.00 &     1.00 &     1.00 &     1.00 &     1.00 &     1.00\\ & 4 &     1.00 &     0.87 &     1.00 &     1.00 &     1.00 &     1.00 &     1.00 &     1.00\\
 \hline 
 \multirow{5}{*}{5} & 1  &     1.00 &     0.86 &     1.00 &     1.00 &     1.00 &     1.00 &     1.00 &     1.00\\ & 2 &     1.00 &     0.83 &     1.00 &     1.00 &     1.00 &     1.00 &     1.00 &     1.00\\ & 3 &     1.00 &     0.79 &     1.00 &     1.00 &     1.00 &     1.00 &     1.00 &     1.00\\ & 4 &     1.00 &     0.86 &     1.00 &     1.00 &     1.00 &     1.00 &     1.00 &     1.00\\ & 5 &     1.00 &     0.73 &     1.00 &     1.00 &     1.00 &     1.00 &     1.00 &     1.00\\
 \hline 
 \end{tabular}

        }
    \end{center}
    Note that the variable 1, and 2 here just denote the first and second relevant variables of each class, and they are not the same among classes.
\end{table}

\section{Real Data Examples}
\subsection{Anuran Species Classification}

Anuran calls data set  was used in several classification tasks related to the challenge of anuran species recognition through their calls \citep[see][]{Dua:2017}, 
and was created by segmenting 60 audio records belonging to 4 different families, 8 genus, and 10 species, and each audio corresponds to one specimen.
There are 7195 syllables identified from the 60 bioacoustics signals after segmenting, and each syllable is represented by a set of variables extracted by Mel-Frequency Spectral Coefficients (MFCCs), which performs a spectral analysis based on a triangular filter-bank logarithmically spaced in the frequency domain. 
Therefore, each sample in this data set is denoted by a variable set of MFCCS coefficients belonging to a special species. 
We focus on classification of the main 7 out of 10 species: Leptodactylus fuscus, Adenomera andreae, Adenomera hylaedactyla, Hyla minuta, Hypsiboas cinerascens, Hypsiboas cordobae, and  Ameerega trivittata. 
In addition to the original dataset, we add 5 normally distributed noise attributes, with mean 0 and variance 1, as the irrelevant variables. 
{\color{black}In each run, we randomly select 100 and 50 examples from each species as the training and testing (evaluation) data, respectively, 
and report the results are based on 100 replications.}

\begin{figure}[t]
\begin{center}
	\includegraphics[width=0.8\textwidth]{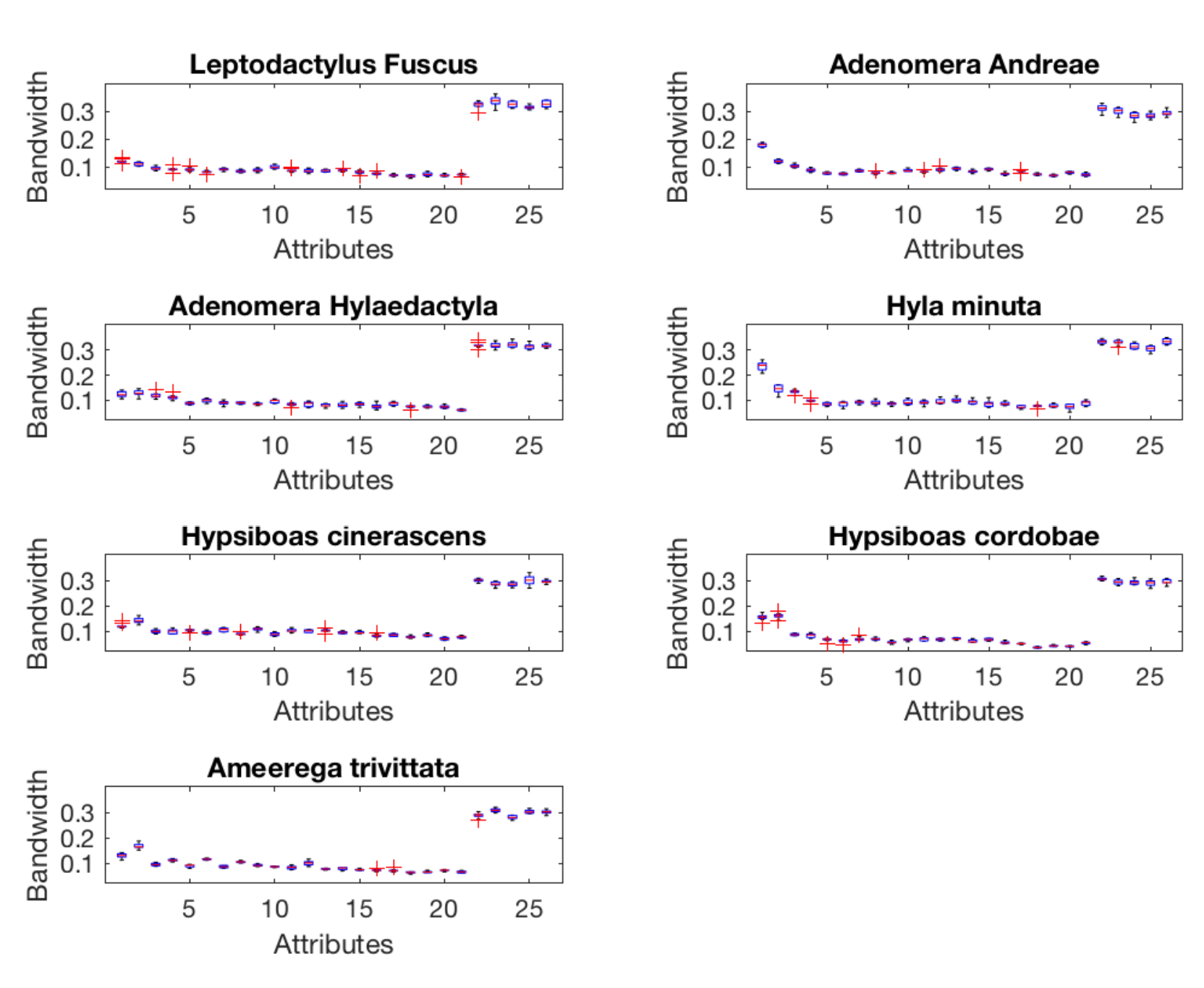}
	\caption{The box-plots of mean predicted bandwidths for anuran calls.}
	\label{fig:frogs2}
\end{center}
\end{figure}

Figure \ref{fig:frogs2} shows the box-plots of the mean predicted bandwidths with noise variables based on 100 trials, 
where the bandwidths of the last 5 added noise variables remain large, and the other bandwidths shrink as expected. 
This result indicates that the last 5 variables are irrelevant, which is consistent with our setup,
and the proposed method successfully detects the relevant variables in this example.
Table \ref{tb:frogs1} summarizes the classification performance of the proposed NPKDC-vd method 
with both the original data and the data with 5 additional irrelevant variables.  
The results with the original data are used as our comparison baseline.  
Note that there are 3 popularly used performance indexes: accuracy, specificity, and precision, in this table.
The results of NPKDC-vd show about $3\%$ drop in accuracy and precision when there are irrelevant variables in the data set,
and specificity remains similar.
The differences in results of NPKDC-vd between these two data sets are not statistically significant.
Table \ref{SVMExAnuran} shows results of SVM and SVMfs under a similar setup.  
If we are only interested in classification accuracy, then SVMfs is the best choice in this example.  
However, SVMfs is built on multiple binary classifiers, and cannot have good interpretation ability. 
Thus, it provides less information about each class.

We summarize the means of Z-scores of the predicted bandwidths for all variables in Table \ref{Table:Frog_Z}.
Most of the variables are identified.  However, we found that besides the added irrelevant variables, some original variables such as Variable 1 and 2 in Hypsiboas Cordobae, and Ameerega Trivittata species are also regarded as irrelevant, which are new findings, and were not reported in the literature before.

\begin{table}\fontsize{8}{7}\selectfont
\caption{Classification results of anuran species.}
\label{tb:frogs1-all}

\begin{subtable}{0.9\linewidth}
\centering

\caption{Classification results  of NPKDC-vd}
\label{tb:frogs1}

\begin{tabular}{l|r|r|r|r|r|r}
\hline
     & \multicolumn{3} {c|}{Original dataset} & \multicolumn{3} {c}{Extended dataset}\\
     & \multicolumn{3} {c|}{(without noise variables)} & \multicolumn{3} {c}{(with noise variables)}\\ \hline

     & Accuracy & Precision & Specificity  & Accuracy & Precision & Specificity\\ \hline
mean & 0.9155   & 0.9173      & 0.9859    & 0.8741   & 0.8807      & 0.9790\\
Std  & 0.0124   & 0.0019      & 0.0021   & 0.0142   & 0.0135      & 0.0024\\ \hline
\end{tabular}

   \caption{Classification results of SVM} 
   \label{SVMExAnuran}
   
	\scalebox{0.9}{
        \begin{tabular}{l|r|r|r|r|r|r}
	\hline
     & \multicolumn{3} {c|}{Original dataset} & \multicolumn{3} {c}{Extended dataset}\\
     & \multicolumn{3} {c|}{(without noise features)} & \multicolumn{3} {c}{(with noise features)}\\ \hline

     & Accuracy & Precision & Specificity  & Accuracy & Precision & Specificity\\ \hline
mean & 0.7606   & 0.9071      & 0.9734    & 0.7488   & 0.9081      & 0.9721\\
Std  & 0.0806   & 0.0118      & 0.0090   & 0.0895   & 0.0132      & 0.0099\\ \hline
\end{tabular}

        }
      \end{subtable}
\end{table}

\begin{table}[t]\fontsize{8}{7}\selectfont
\begin{center}
    \caption{Anuran Calls: The means of Z-scores of the mean predicted bandwidths for all attributes, where the cells with gray background are {\bf relevant} variables.}
    \label{Table:Frog_Z}
\begin{tabular}{l|ccccccccc}\hline 
\diagbox{Species~}{variable~~}& 1 & 2 & 3 & 4 & 5 & 6 & 7 & 8  & 9  \\ \hline \hline 
 Leptodactylus Fuscus &    -0.14\cellcolor[gray]{0.9}  &    -0.27\cellcolor[gray]{0.9}  &    -0.38\cellcolor[gray]{0.9}  &    -0.44\cellcolor[gray]{0.9}  &    -0.37\cellcolor[gray]{0.9}  &    -0.47\cellcolor[gray]{0.9}  &    -0.43\cellcolor[gray]{0.9} &    -0.57\cellcolor[gray]{0.9}  &    -0.44\cellcolor[gray]{0.9}  \\ 
 Adenomera Andreae &     0.53  &    -0.04\cellcolor[gray]{0.9}  &    -0.33\cellcolor[gray]{0.9}  &    -0.39\cellcolor[gray]{0.9}  &    -0.5\cellcolor[gray]{0.9}9  &    -0.59\cellcolor[gray]{0.9}  &    -0.45\cellcolor[gray]{0.9} &    -0.59\cellcolor[gray]{0.9}  &    -0.55\cellcolor[gray]{0.9}  \\ 
 Adenomera Hylaedactyla &    -0.24\cellcolor[gray]{0.9}  &    -0.27\cellcolor[gray]{0.9}  &    -0.28\cellcolor[gray]{0.9}  &    -0.29\cellcolor[gray]{0.9}  &    -0.53\cellcolor[gray]{0.9}  &    -0.24\cellcolor[gray]{0.9}  &    -0.48\cellcolor[gray]{0.9} &    -0.46\cellcolor[gray]{0.9}  &    -0.55\cellcolor[gray]{0.9}  \\ 
 Hyla Minuta &     0.80  &    -0.10\cellcolor[gray]{0.9}  &     0.01  &    -0.43\cellcolor[gray]{0.9}  &    -0.64\cellcolor[gray]{0.9}  &    -0.58\cellcolor[gray]{0.9}  &    -0.62\cellcolor[gray]{0.9} &    -0.51\cellcolor[gray]{0.9}  &    -0.60\cellcolor[gray]{0.9}  \\ 
 Hypsiboas Cinerascens &    -0.11\cellcolor[gray]{0.9}  &     0.06  &    -0.31\cellcolor[gray]{0.9}  &    -0.40\cellcolor[gray]{0.9}  &    -0.42\cellcolor[gray]{0.9}  &    -0.55\cellcolor[gray]{0.9}  &    -0.23\cellcolor[gray]{0.9} &    -0.57\cellcolor[gray]{0.9}  &    -0.40\cellcolor[gray]{0.9}  \\ 
 Hypsiboas Cordobae &     0.29  &     0.53  &    -0.26\cellcolor[gray]{0.9}  &    -0.35\cellcolor[gray]{0.9}  &    -0.44\cellcolor[gray]{0.9}  &    -0.57\cellcolor[gray]{0.9}  &    -0.40\cellcolor[gray]{0.9} &    -0.56\cellcolor[gray]{0.9}  &    -0.58\cellcolor[gray]{0.9}  \\ 
 Ameerega Trivittata &     0.08  &     0.20  &    -0.28\cellcolor[gray]{0.9}  &    -0.40\cellcolor[gray]{0.9}  &    -0.50\cellcolor[gray]{0.9}  &    -0.22\cellcolor[gray]{0.9}  &    -0.55\cellcolor[gray]{0.9} &    -0.37\cellcolor[gray]{0.9}  &    -0.36\cellcolor[gray]{0.9}  \\ 
\hline \hline
\diagbox{Species~}{variable~~} & 10 & 11 & 12 & 13 & 14 & 15 & 16 & 17  & 18 \\ \hline \hline
 Leptodactylus Fuscus &     -0.30\cellcolor[gray]{0.9} &    -0.43\cellcolor[gray]{0.9} &    -0.41\cellcolor[gray]{0.9} &    -0.51\cellcolor[gray]{0.9} &    -0.45\cellcolor[gray]{0.9} &    -0.55\cellcolor[gray]{0.9} &    -0.62\cellcolor[gray]{0.9} &    -0.64\cellcolor[gray]{0.9} &    -0.62\cellcolor[gray]{0.9}  \\ 
 Adenomera Andreae &     -0.47\cellcolor[gray]{0.9} &    -0.50\cellcolor[gray]{0.9} &    -0.43\cellcolor[gray]{0.9} &    -0.46\cellcolor[gray]{0.9} &    -0.55\cellcolor[gray]{0.9} &    -0.47\cellcolor[gray]{0.9} &    -0.58\cellcolor[gray]{0.9} &    -0.59\cellcolor[gray]{0.9} &    -0.67\cellcolor[gray]{0.9}  \\ 
 Adenomera Hylaedactyla &     -0.41\cellcolor[gray]{0.9} &    -0.55\cellcolor[gray]{0.9} &    -0.45\cellcolor[gray]{0.9} &    -0.48\cellcolor[gray]{0.9} &    -0.49\cellcolor[gray]{0.9} &    -0.49\cellcolor[gray]{0.9} &    -0.58\cellcolor[gray]{0.9} &    -0.61\cellcolor[gray]{0.9} &    -0.64\cellcolor[gray]{0.9}  \\ 
 Hyla minuta &     -0.50\cellcolor[gray]{0.9} &    -0.47\cellcolor[gray]{0.9} &    -0.49\cellcolor[gray]{0.9} &    -0.45\cellcolor[gray]{0.9} &    -0.50\cellcolor[gray]{0.9} &    -0.54\cellcolor[gray]{0.9} &    -0.56\cellcolor[gray]{0.9} &    -0.62\cellcolor[gray]{0.9} &    -0.72\cellcolor[gray]{0.9}  \\ 
 Hypsiboas Cinerascens &     -0.59\cellcolor[gray]{0.9} &    -0.53\cellcolor[gray]{0.9} &    -0.46\cellcolor[gray]{0.9} &    -0.49\cellcolor[gray]{0.9} &    -0.50\cellcolor[gray]{0.9} &    -0.46\cellcolor[gray]{0.9} &    -0.73\cellcolor[gray]{0.9} &    -0.56\cellcolor[gray]{0.9} &    -0.65\cellcolor[gray]{0.9}  \\ 
 Hypsiboas Cordobae &     -0.55\cellcolor[gray]{0.9} &    -0.49\cellcolor[gray]{0.9} &    -0.47\cellcolor[gray]{0.9} &    -0.51\cellcolor[gray]{0.9} &    -0.48\cellcolor[gray]{0.9} &    -0.55\cellcolor[gray]{0.9} &    -0.68\cellcolor[gray]{0.9} &    -0.68\cellcolor[gray]{0.9} &    -0.71\cellcolor[gray]{0.9}  \\ 
 Ameerega Trivittata &     -0.49\cellcolor[gray]{0.9} &    -0.52\cellcolor[gray]{0.9} &    -0.39\cellcolor[gray]{0.9} &    -0.63\cellcolor[gray]{0.9} &    -0.48\cellcolor[gray]{0.9} &    -0.65\cellcolor[gray]{0.9} &    -0.65\cellcolor[gray]{0.9} &    -0.64\cellcolor[gray]{0.9} &    -0.71\cellcolor[gray]{0.9}  \\ 
\hline \hline
\diagbox{Species~}{variable~~} & 19 & 20 & 21 & 22 & 23 & 24 & 25 & 26  &  \\ \hline \hline
 Leptodactylus Fuscus &     -0.59\cellcolor[gray]{0.9} &    -0.65\cellcolor[gray]{0.9} &    -0.66\cellcolor[gray]{0.9} &     2.09 &     1.81 &     2.10 &     1.98 &     1.96 &   \\ 
 Adenomera Andreae &     -0.72\cellcolor[gray]{0.9} &    -0.63\cellcolor[gray]{0.9} &    -0.65\cellcolor[gray]{0.9} &     2.03 &     1.92 &     1.93 &     1.99 &     1.85 &   \\ 
 Adenomera Hylaedactyla &     -0.63\cellcolor[gray]{0.9} &    -0.60\cellcolor[gray]{0.9} &    -0.67\cellcolor[gray]{0.9} &     2.02 &     1.87 &     2.08 &     2.06 &     1.90 &   \\ 
 Hyla Minuta &     -0.71\cellcolor[gray]{0.9} &    -0.69\cellcolor[gray]{0.9} &    -0.62\cellcolor[gray]{0.9} &     1.90 &     1.95 &     1.83 &     2.01 &     1.85 &   \\ 
 Hypsiboas Cinerascens &     -0.60\cellcolor[gray]{0.9} &    -0.70\cellcolor[gray]{0.9} &    -0.61\cellcolor[gray]{0.9} &     1.90 &     2.15 &     1.92 &     1.84 &     2.04 &   \\ 
 Hypsiboas Cordobae &     -0.73\cellcolor[gray]{0.9} &    -0.75\cellcolor[gray]{0.9} &    -0.65\cellcolor[gray]{0.9} &     1.96 &     2.06 &     1.78 &     1.92 &     1.89 &   \\ 
 Ameerega Trivittata &     -0.69\cellcolor[gray]{0.9} &    -0.74\cellcolor[gray]{0.9} &    -0.79\cellcolor[gray]{0.9} &     1.95 &     1.97 &     2.06 &     1.93 &     1.84 &   \\ 
\hline \hline
\end{tabular}
\end{center}
\end{table}

\begin{table}    
        \begin{center}
          \caption{ The probabilities of relevant variable identified by sequential variable selection, based on 1000 iterations, where the cells with gray background are variables with probability larger than 0.5}
        \label{Anuranfs}
	\scalebox{0.8}{
\begin{tabular}{c|cccccccccc}\hline 
\diagbox{group~}{variable~~}& 1 & 2 & 3 & 4 & 5 & 6 & 7 & 8  & 9 & 10  \\ \hline \hline 
 1$\sim$10 & 0.1100\cellcolor[gray]{0.9} & 0.2000\cellcolor[gray]{0.9} & 0.2200\cellcolor[gray]{0.9} & 0.2900\cellcolor[gray]{0.9} & 0.3900\cellcolor[gray]{0.9} & 0.5000\cellcolor[gray]{0.9} &     0.4900\cellcolor[gray]{0.9} & 0.4100\cellcolor[gray]{0.9}  & 0.4400\cellcolor[gray]{0.9} & 0.3400\cellcolor[gray]{0.9}  \\ 

\hline \hline\diagbox{group~}{variable~~} & 11 & 12 & 13 & 14 & 15 & 16 & 17 & 18  & 19 & 20  \\ \hline \hline
 1$\sim$10 & 0.2700\cellcolor[gray]{0.9} & 0.1300\cellcolor[gray]{0.9} & 0.0900\cellcolor[gray]{0.9} & 0.0300\cellcolor[gray]{0.9} & 0.0100\cellcolor[gray]{0.9} & 0.0000 & 0.0000 & 0.0000  & 0.0000 & 0.0000  \\ 

\hline \hline\diagbox{group~}{variable~~} & 21 & 22 & 23 & 24 & 25 & 26 & 27 & 28  & 29 & 30  \\ \hline \hline
 1$\sim$10 & 0.0000 & 0.0000 & 0.0000 & 0.0000 & 0.0000 & 0.0000 &     0.0000 & 0.0000  & 0.0000 & 0.0000  \\ 
\hline \hline
\end{tabular}
        }
    \end{center}
\end{table}

\subsection{Handwritten digit dataset}
We use a subset from the MINST database of handwritten digits, 
where digits have been size-normalized and centered on a fixed-size image. 
We randomly select 100 gray images from this data set, for training, and another 100 images for testing from this database. 
(For the details about this database, please refer to \cite{lecun2010mnist}.)

We resize each handwritten digit image to $64 = 8\times 8$ pixels, and treat each pixel as a variable. 
It becomes a 10-class classification problem through 64-dimensional density estimation when applying the proposed method. 
Table \ref{tb:digit1} shows the classification results of NPKDC-vd, where accuracy, precision, and specificity for all digits are greater than $95\%$,
Table \ref{tb:digit2} shows the results of both SVM and SVMfs. 
From Table \ref{tb:digit1-all}, we found that NPKDC-vd largely outperforms these two SVM-based methods in all three measurements, 
since the relations among classes in this example are rather complicated, and binary-classifier-based methods will suffer from complicated ``voting-like'' schemes.

Table \ref{Table_Digits} shows the means of Z-scores of the mean predicted bandwidths for all 64 attributes/pixels.
The corresponding bandwidths of the relevant pixels could drop to a very small value.
Moreover, because the background pixels of the image data have a density close to a point mass, instead of a uniform distribution,  
the variables with large interquartile ranges (IQR) are irrelevant. 
Figures of the box-plots for the mean of the selected bandwidths of the testing images based on 100 trials are in Supplementary, 
which provides similar information as in Table \ref{Table_Digits}.
The bandwidths of variables in this situation are the pixels on the edges of the image, such as 1, 8, 9, 16, 17, 24, 25, 32, 33, 40, 41, 48, 49, 56, 57, and 64, 
and their IQRs  are usually large (see also Figures S.1 and S.2 in Supplementary). 
Table \ref{tb:digitprobSVM} shows the probabilities of relevant variable identification by SVM with sequential variable selection, based on 1000 iterations, for handwritten digits.  
We can see that the pixels/variables identified via the sequential variable selection of SVMfs scatter around all pixels, and cannot focus on some particular variables for each class (digit).  Thus, in this case, SVMfs cannot provide neither stable information about each class, nor satisfactory classification performance.

\begin{table}[t]\fontsize{9}{8}\selectfont
\centering

\caption{Classification results of handwritten digit dataset.}
\label{tb:digit1-all}

\begin{subtable}{0.9\linewidth}
\centering

\caption{Classification results of NPKDC-vd.}
\label{tb:digit1}

\begin{tabular}{l|r|r|r}
\hline
     & Accuracy & Precision & Specificity   \\ \hline\hline
mean & 0.9698   & 0.9966      & 0.9797  \\
Std  & 0.0047   & 0.0005      & 0.0030   \\ \hline
\end{tabular}

        \caption{Classification results of SVM and SVMfs}
        \label{tb:digit2}
	\scalebox{1}{
 \begin{tabular}{l|r|r|r|r|r|r}
\hline
     & \multicolumn{3} {c|}{SVM} & \multicolumn{3} {c}{SVMfs (with feature selection)}\\
\hline

     & Accuracy & Precision & Specificity  & Accuracy & Precision & Specificity\\ \hline
mean & 0.2739   & 0.8721      & 0.9193    & 0.7606   & 0.9071      & 0.9734\\
Std  & 0.0227   & 0.0303      & 0.0025   & 0.0806   & 0.0118      & 0.0090\\ \hline
\end{tabular}
        }
\end{subtable}
\end{table}

\begin{table}[ht!]\fontsize{6}{5}\selectfont
\begin{center}      
    \caption{\small Handwritten Digits: The mean Z-scores of the mean predicted bandwidths for all attributes. The cells with {\bf gray background are irrelevant variables} for its corresponding digit.}
    \label{Table_Digits}
\begin{tabular}{c|ccccccccccccc}\hline \hline
\multirow{2}{*}{Digits} & \multicolumn{13}{c}{Attribute}\\
& 1 & 2 & 3 & 4 & 5 & 6 & 7 & 8  & 9 & 10 & 11 & 12  & 13 \\ \hline \hline 
 1 &    -0.80\cellcolor{gray!10}  &    -0.77\cellcolor{gray!10}  &     0.81  &     1.07  &     1.17  &     1.12  &    -0.40\cellcolor{gray!10} &    -0.80\cellcolor{gray!10}  &    -0.80\cellcolor{gray!10} &    -0.33\cellcolor{gray!10}  &     0.90 &     0.02  &    -0.38\cellcolor{gray!10} \\ 
 2 &    -1.18\cellcolor{gray!10}  &     0.04  &     1.05  &     0.71  &     0.95  &    -0.82\cellcolor{gray!10}  &    -1.16\cellcolor{gray!10} &    -1.18\cellcolor{gray!10}  &    -1.18\cellcolor{gray!10} &     1.12  &     0.40 &     0.77  &     0.93 \\ 
 3 &    -1.27\cellcolor{gray!10}  &    -0.12\cellcolor{gray!10}  &     0.98  &     0.50  &     0.88  &     0.91  &    -0.98\cellcolor{gray!10} &    -1.25\cellcolor{gray!10}  &    -1.27\cellcolor{gray!10} &     1.02  &     0.70 &     0.93  &     0.81 \\ 
 4 &    -0.94\cellcolor{gray!10}  &    -0.93\cellcolor{gray!10}  &    -0.75\cellcolor{gray!10}  &     0.72  &     0.80  &     0.66  &    -0.78\cellcolor{gray!10} &    -0.87\cellcolor{gray!10}  &    -0.94\cellcolor{gray!10} &    -0.87\cellcolor{gray!10}  &     0.26 &     0.65  &     0.75 \\ 
 5 &    -1.22\cellcolor{gray!10}  &    -0.17\cellcolor{gray!10}  &     0.97  &     0.91  &     0.80  &     0.92  &     0.94 &    -1.17\cellcolor{gray!10}  &    -1.22\cellcolor{gray!10} &     0.75  &     0.18 &     0.82  &     0.69 \\ 
 6 &    -1.15\cellcolor{gray!10}  &    -1.15\cellcolor{gray!10}  &     0.32  &     1.02  &     1.11  &    -0.34\cellcolor{gray!10}  &    -1.15\cellcolor{gray!10} &    -1.15\cellcolor{gray!10}  &    -1.15\cellcolor{gray!10} &    -1.06\cellcolor{gray!10}  &     1.14 &     0.47  &     0.95 \\ 
 7 &    -1.11\cellcolor{gray!10}  &    -0.91\cellcolor{gray!10}  &     1.07  &     0.87  &     0.46  &     0.78  &     1.13 &    -0.36\cellcolor{gray!10}  &    -1.11\cellcolor{gray!10} &    -0.47\cellcolor{gray!10}  &     1.19 &     1.09  &     1.03 \\ 
 8 &    -1.28\cellcolor{gray!10}  &    -1.04\cellcolor{gray!10}  &     0.80  &     0.61  &     0.69  &     0.89  &    -0.70\cellcolor{gray!10} &    -1.28\cellcolor{gray!10}  &    -1.27\cellcolor{gray!10} &     0.73  &     0.50 &     0.85  &     0.85 \\ 
 9 &    -1.26\cellcolor{gray!10}  &    -0.86\cellcolor{gray!10}  &     0.86  &     0.81  &     0.85  &     0.69  &    -0.73\cellcolor{gray!10} &    -1.21\cellcolor{gray!10}  &    -1.26\cellcolor{gray!10} &     0.64  &     0.35 &     0.86  &     0.78 \\ 
 10 &    -1.27\cellcolor{gray!10}  &    -1.26\cellcolor{gray!10}  &     0.91  &     0.74  &     1.11  &     0.39  &    -1.25\cellcolor{gray!10} &    -1.27\cellcolor{gray!10}  &    -1.27\cellcolor{gray!10} &     0.13  &     0.87 &     1.02  &     0.99 \\ \hline \hline
 \multirow{2}{*}{Digits} & \multicolumn{13}{c}{Attribute}\\
 & 14 & 15 & 16 & 17 & 18 & 19 & 20 & 21  & 22 & 23 & 24 & 25  & 26 \\ \hline \hline 
 1 &     1.20  &    -0.22\cellcolor{gray!10}  &    -0.80\cellcolor{gray!10}  &    -0.79\cellcolor{gray!10}  &    -0.22\cellcolor{gray!10}  &     1.23  &    -0.23\cellcolor{gray!10} &    -0.65\cellcolor{gray!10}  &     1.15 &    -0.43\cellcolor{gray!10}  &    -0.80\cellcolor{gray!10} &    -0.78\cellcolor{gray!10}  &     0.01 \\ 
 2 &     0.27  &    -1.13\cellcolor{gray!10}  &    -1.18\cellcolor{gray!10}  &    -1.18\cellcolor{gray!10}  &     1.03  &     1.17  &     0.96 &     0.90  &     0.49 &    -1.09\cellcolor{gray!10}  &    -1.18\cellcolor{gray!10} &    -1.18\cellcolor{gray!10}  &     0.05 \\ 
 3 &     0.99  &    -0.51\cellcolor{gray!10}  &    -1.26\cellcolor{gray!10}  &    -1.27\cellcolor{gray!10}  &     0.49  &     0.85  &     0.78 &     0.92  &     1.05 &    -0.71\cellcolor{gray!10}  &    -1.27\cellcolor{gray!10} &    -1.27\cellcolor{gray!10}  &    -0.98\cellcolor{gray!10} \\ 
 4 &     0.05  &     0.01  &    -0.64\cellcolor{gray!10}  &    -0.94\cellcolor{gray!10}  &    -0.56\cellcolor{gray!10}  &     0.73  &     0.76 &     0.57  &     0.69 &     0.90  &    -0.59\cellcolor{gray!10} &    -0.94\cellcolor{gray!10}  &     0.42 \\ 
 5 &     0.90  &     0.62  &    -1.19\cellcolor{gray!10}  &    -1.20\cellcolor{gray!10}  &     0.89  &     0.33  &     0.84 &     0.53  &    -0.31\cellcolor{gray!10} &    -1.14\cellcolor{gray!10}  &    -1.22\cellcolor{gray!10} &    -1.22\cellcolor{gray!10}  &     0.88 \\ 
 6 &    -0.49\cellcolor{gray!10}  &    -1.15\cellcolor{gray!10}  &    -1.15\cellcolor{gray!10}  &    -1.15\cellcolor{gray!10}  &    -0.14\cellcolor{gray!10}  &     0.90  &     1.06 &    -0.81\cellcolor{gray!10}  &    -1.14\cellcolor{gray!10} &    -1.15\cellcolor{gray!10}  &    -1.15\cellcolor{gray!10} &    -1.15\cellcolor{gray!10}  &     0.70 \\ 
 7 &     0.66  &     0.96  &    -0.55\cellcolor{gray!10}  &    -1.11\cellcolor{gray!10}  &    -0.86\cellcolor{gray!10}  &    -0.21\cellcolor{gray!10}  &    -0.56\cellcolor{gray!10} &     0.90  &     0.80 &     0.86  &    -0.96\cellcolor{gray!10} &    -1.11\cellcolor{gray!10}  &    -0.67\cellcolor{gray!10} \\ 
 8 &     0.73  &     0.60  &    -1.28\cellcolor{gray!10}  &    -1.27\cellcolor{gray!10}  &     0.88  &     0.70  &     0.76 &     0.88  &     0.71 &     0.54  &    -1.28\cellcolor{gray!10} &    -1.28\cellcolor{gray!10}  &     0.33 \\ 
 9 &     0.97  &     0.13  &    -1.13\cellcolor{gray!10}  &    -1.26\cellcolor{gray!10}  &     0.90  &     0.73  &     0.82 &     0.89  &     0.70 &     0.51  &    -1.14\cellcolor{gray!10} &    -1.26\cellcolor{gray!10}  &     0.59 \\ 
 10 &     1.06  &    -0.81\cellcolor{gray!10}  &    -1.2\cellcolor{gray!10}7  &    -1.27\cellcolor{gray!10}  &     1.03  &     0.38  &     0.94 &     0.52  &     0.78 &     0.73  &    -1.27\cellcolor{gray!10} &    -1.27\cellcolor{gray!10}  &     0.86 \\ \hline \hline
 \multirow{2}{*}{Digits} & \multicolumn{13}{c}{Attribute}\\
 & 27 & 28 & 29 & 30 & 31 & 32 & 33 & 34  & 35 & 36 & 37 & 38  & 39 \\ \hline \hline 
 1 &     1.23  &    -0.43\cellcolor{gray!10}  &    -0.61\cellcolor{gray!10}  &     1.13  &    -0.48\cellcolor{gray!10}  &    -0.80\cellcolor{gray!10}  &    -0.80\cellcolor{gray!10} &     0.02  &     1.26 &    -0.32\cellcolor{gray!10}  &    -0.25\cellcolor{gray!10} &     1.26  &    -0.65\cellcolor{gray!10} \\ 
 2 &     0.49  &     0.76  &     0.94  &     0.50  &    -1.17\cellcolor{gray!10}  &    -1.18\cellcolor{gray!10}  &    -1.18\cellcolor{gray!10} &    -1.09\cellcolor{gray!10}  &    -0.62\cellcolor{gray!10} &     0.99  &     0.86 &    -0.01\cellcolor{gray!10}  &    -1.17\cellcolor{gray!10} \\ 
 3 &     0.44  &     1.02  &    -0.01\cellcolor{gray!10}  &     0.98  &    -1.00\cellcolor{gray!10}  &    -1.27\cellcolor{gray!10}  &    -1.27\cellcolor{gray!10} &    -1.17\cellcolor{gray!10}  &     0.17 &     1.02  &     0.86 &     0.68  &     0.73 \\ 
 4 &     0.27  &     0.93  &     0.49  &     0.70  &     0.83  &    -0.83\cellcolor{gray!10}  &    -0.93\cellcolor{gray!10} &     0.80  &    -0.32\cellcolor{gray!10} &     0.90  &     0.83 &    -0.37\cellcolor{gray!10}  &     0.92 \\ 
 5 &    -0.16\cellcolor{gray!10}  &     0.53  &     0.58  &     0.54  &    -0.58\cellcolor{gray!10}  &    -1.22\cellcolor{gray!10}  &    -1.22\cellcolor{gray!10} &     0.39  &     0.99 &     0.69  &     0.83 &     0.87  &     0.20 \\ 
 6 &     0.55  &     1.17  &     0.31  &    -0.21\cellcolor{gray!10}  &    -1.12\cellcolor{gray!10}  &    -1.15\cellcolor{gray!10}  &    -1.15\cellcolor{gray!10} &     0.91  &     0.30 &     0.92  &     0.93 &     1.05  &     0.34 \\ 
 7 &     0.86  &     1.17  &     0.81  &     0.65  &     1.02  &    -1.08\cellcolor{gray!10}  &    -1.11\cellcolor{gray!10} &     0.06  &     1.13 &     0.37  &    -0.29\cellcolor{gray!10} &     0.82  &     1.15 \\ 
 8 &     0.76  &     0.43  &     0.30  &     0.81  &    -0.40\cellcolor{gray!10}  &    -1.28\cellcolor{gray!10}  &    -1.28\cellcolor{gray!10} &    -0.75\cellcolor{gray!10}  &     0.85 &     0.15  &     0.60 &     0.39  &    -0.57\cellcolor{gray!10} \\ 
 9 &     0.77  &     0.83  &     0.84  &     0.08  &     0.94  &    -1.26\cellcolor{gray!10}  &    -1.26\cellcolor{gray!10} &    -1.13\cellcolor{gray!10}  &     0.36 &     0.77  &     0.90 &     0.70  &     0.82 \\ 
 10 &     0.98  &     0.02  &    -1.14\cellcolor{gray!10}  &     1.09  &     0.49  &    -1.27\cellcolor{gray!10}  &    -1.27\cellcolor{gray!10} &     0.88  &     1.06 &    -0.72\cellcolor{gray!10}  &    -1.14\cellcolor{gray!10} &     1.10  &     0.43 \\ \hline \hline
 \multirow{2}{*}{Digits} & \multicolumn{13}{c}{Attribute}\\
 & 40 & 41 & 42 & 43 & 44 & 45 & 46 & 47  & 48 & 49 & 50 & 51  & 52 \\ \hline \hline 
 1 &    -0.80\cellcolor{gray!10}  &    -0.80\cellcolor{gray!10}  &    -0.30\cellcolor{gray!10}  &     1.05  &    -0.20\cellcolor{gray!10}  &    -0.28\cellcolor{gray!10}  &     1.17 &    -0.64\cellcolor{gray!10}  &    -0.80\cellcolor{gray!10} &    -0.80\cellcolor{gray!10}  &    -0.64\cellcolor{gray!10} &     0.89  &    -0.32\cellcolor{gray!10} \\ 
 2 &    -1.18\cellcolor{gray!10}  &    -1.18\cellcolor{gray!10}  &    -1.01\cellcolor{gray!10}  &     0.52  &     0.87  &     1.02  &    -0.22\cellcolor{gray!10} &    -0.74\cellcolor{gray!10}  &    -1.18\cellcolor{gray!10} &    -1.16\cellcolor{gray!10}  &     0.11 &     0.95  &    -0.31\cellcolor{gray!10} \\ 
 3 &    -1.27\cellcolor{gray!10}  &    -1.27\cellcolor{gray!10}  &    -1.25\cellcolor{gray!10}  &    -0.39\cellcolor{gray!10}  &    -0.76\cellcolor{gray!10}  &     0.22  &     0.95 &     0.84  &    -1.26\cellcolor{gray!10} &    -1.27\cellcolor{gray!10}  &     0.26 &     1.00  &     1.00 \\ 
 4 &    -0.94\cellcolor{gray!10}  &    -0.46\cellcolor{gray!10}  &     0.72  &     0.44  &     0.72  &    -0.23\cellcolor{gray!10}  &     0.55 &     0.42  &    -0.94\cellcolor{gray!10} &    -0.74\cellcolor{gray!10}  &    -0.10\cellcolor{gray!10} &     0.19  &     0.80 \\ 
 5 &    -1.22\cellcolor{gray!10}  &    -1.22\cellcolor{gray!10}  &    -1.14\cellcolor{gray!10}  &    -0.87\cellcolor{gray!10}  &    -0.74\cellcolor{gray!10}  &     0.48  &     0.79 &     0.36  &    -1.21\cellcolor{gray!10} &    -1.22  &     0.02 &     1.01  &     0.94 \\ 
 6 &    -1.15\cellcolor{gray!10}  &    -1.15\cellcolor{gray!10}  &     0.79  &     0.05  &     1.08  &     1.02  &     1.04 &     1.12  &    -0.92\cellcolor{gray!10} &    -1.15\cellcolor{gray!10}  &    -0.55\cellcolor{gray!10} &     1.05  &     0.97 \\ 
 7 &    -1.11\cellcolor{gray!10}  &    -1.11\cellcolor{gray!10}  &    -0.42\cellcolor{gray!10}  &     0.62  &     0.86  &     0.93  &     0.72 &    -0.41\cellcolor{gray!10}  &    -1.11\cellcolor{gray!10} &    -1.11\cellcolor{gray!10}  &    -1.07\cellcolor{gray!10} &     0.79  &     0.53 \\ 
 8 &    -1.28\cellcolor{gray!10}  &    -1.28\cellcolor{gray!10}  &    -0.12\cellcolor{gray!10}  &     0.83  &     0.89  &     0.77  &     0.81 &    -0.13\cellcolor{gray!10}  &    -1.28\cellcolor{gray!10} &    -1.28\cellcolor{gray!10}  &    -0.25\cellcolor{gray!10} &     0.77  &     0.88 \\ 
 9 &    -1.26\cellcolor{gray!10}  &    -1.26\cellcolor{gray!10}  &    -1.16\cellcolor{gray!10}  &    -0.92\cellcolor{gray!10}  &    -0.97\cellcolor{gray!10}  &    -0.15\cellcolor{gray!10}  &     0.96 &     0.70  &    -1.25\cellcolor{gray!10} &    -1.26\cellcolor{gray!10}  &    -0.47\cellcolor{gray!10} &     0.66  &     0.94 \\ 
 10 &    -1.27\cellcolor{gray!10}  &    -1.27\cellcolor{gray!10}  &     0.75  &     0.74  &    -0.00\cellcolor{gray!10}  &    -0.26\cellcolor{gray!10}  &     1.00 &     0.96  &    -1.27\cellcolor{gray!10} &    -1.27\cellcolor{gray!10}  &    -0.24\cellcolor{gray!10} &     0.79  &     1.12 \\ \hline \hline
 \multirow{2}{*}{Digits} & \multicolumn{13}{c}{Attribute}\\
 & 53 & 54 & 55 & 56 & 57 & 58 & 59 & 60  & 61 & 62 & 63 & 64  &  \\ \hline \hline 
 1 &    -0.30\cellcolor{gray!10}  &     1.18  &     0.03  &    -0.79\cellcolor{gray!10}  &    -0.80\cellcolor{gray!10}  &    -0.78\cellcolor{gray!10}  &     0.83 &     0.94  &     1.18 &     1.13  &     0.59 &    -0.79\cellcolor{gray!10}  \\ 
 2 &     1.02  &     1.12  &     1.11  &    -0.35\cellcolor{gray!10}  &    -1.17\cellcolor{gray!10}  &     0.05  &     1.06 &     0.82  &     0.95 &     1.01  &     1.10 &     0.21  \\ 
 3 &     1.05  &     0.53  &     1.03  &    -1.22\cellcolor{gray!10}  &    -1.27\cellcolor{gray!10}  &    -0.22\cellcolor{gray!10}  &     1.03 &     0.70  &     0.77 &     1.04  &     0.15 &    -1.27\cellcolor{gray!10}  \\ 
 4 &     0.26  &     0.93  &    -0.78\cellcolor{gray!10}  &    -0.94\cellcolor{gray!10}  &    -0.94\cellcolor{gray!10}  &    -0.90\cellcolor{gray!10}  &    -0.76\cellcolor{gray!10} &     0.95  &     0.68 &     0.49  &    -0.94\cellcolor{gray!10} &    -0.94\cellcolor{gray!10}  \\ 
 5 &     0.75  &     0.79  &     0.17  &    -1.21\cellcolor{gray!10}  &    -1.22\cellcolor{gray!10}  &    -0.19\cellcolor{gray!10}  &     0.99 &     0.31  &     0.80 &     0.68  &    -1.01\cellcolor{gray!10} &    -1.21\cellcolor{gray!10}  \\ 
 6 &     1.14  &     1.09  &     1.06  &    -0.43\cellcolor{gray!10}  &    -1.15\cellcolor{gray!10}  &    -1.15\cellcolor{gray!10}  &     0.51 &     1.02  &     0.52 &     1.00  &     1.11 &    -0.86\cellcolor{gray!10}  \\ 
 7 &     0.94  &    -0.99\cellcolor{gray!10}  &    -1.11\cellcolor{gray!10}  &    -1.11\cellcolor{gray!10}  &    -1.11\cellcolor{gray!10}  &    -0.94\cellcolor{gray!10}  &     1.08 &     0.92  &     0.06 &    -1.08\cellcolor{gray!10}  &    -1.11\cellcolor{gray!10} &    -1.11\cellcolor{gray!10}  \\ 
 8 &     0.82  &     0.77  &     0.25  &    -1.28\cellcolor{gray!10}  &    -1.28\cellcolor{gray!10}  &    -1.20\cellcolor{gray!10}  &     0.82 &     0.60  &     0.71 &     0.86  &    -0.53\cellcolor{gray!10} &    -1.27\cellcolor{gray!10}  \\ 
 9 &     0.86  &     0.67  &     0.55  &    -1.23\cellcolor{gray!10}  &    -1.26\cellcolor{gray!10}  &    -0.98\cellcolor{gray!10}  &     0.72 &     0.58  &     0.76 &     0.77  &    -0.09\cellcolor{gray!10} &    -1.26\cellcolor{gray!10}  \\ 
 10 &     1.12  &     0.63  &     0.46  &    -1.27\cellcolor{gray!10}  &    -1.27\cellcolor{gray!10}  &    -1.27\cellcolor{gray!10}  &     0.82 &     0.80  &     0.76 &     1.00  &    -1.04\cellcolor{gray!10} &    -1.27\cellcolor{blue!10}  \\ 
\hline \hline
\end{tabular}
\end{center}
\end{table}

\begin{table}    
        \begin{center}
          \caption{The probabilities of relevant variable identification of SVM with sequential variable selection for handwritten digits based on 1000 runs.}
        \label{tb:digitprobSVM}
	\scalebox{0.8}{
        \begin{tabular}{c|ccccccccccccc}\hline \hline
Attribute& 1 & 2 & 3 & 4 & 5 & 6 & 7 & 8  & 9 & 10 & 11 & 12  & 13 \\ \hline 
 Prob. &    0  &    0.16  &     0.19  &     0.19  &     0.18  &     0.34  &    0.19 &    0.18  &    0.11 &    0.10  &     0.74 &     0.21  &    0.18
  \\ \hline \hline
 Attribute & 14 & 15 & 16 & 17 & 18 & 19 & 20 & 21  & 22 & 23 & 24 & 25  & 26 \\ \hline 
 Prob. &     0.42  &    0.09  &    0.16  &    0.04  &    0.09  &     0.72  &    0.66 &    0.76  &    0.96 &    0.16  &    0.11 &    0  &     0.18 \\ 
\hline \hline
Attribute & 27 & 28 & 29 & 30 & 31 & 32 & 33 & 34  & 35 & 36 & 37 & 38  & 39 \\ \hline 
 Prob. &     0.65  &    0.63  &    0.37  &     0.49  &    0.52  &    0  &    0 &     0.1  &     0.17 &    0.24  &    0.56 &     0.39  &    0.39 \\ 
 \hline \hline
 Attribute & 40 & 41 & 42 & 43 & 44 & 45 & 46 & 47  & 48 & 49 & 50 & 51  & 52 \\ \hline 
 Prob. &    0  &    0.01  &   0.19  &     1  &    0.73  &    0.43  &     0.32 &    0.21  &    0.03 &    0.03  & 0.14  &    0.09 &     0.33\\ 
 \hline \hline
 Attribute & 53 & 54 & 55 & 56 & 57 & 58 & 59 & 60  & 61 & 62 & 63 & 64  &  \\ \hline
 Prob. &   0.41  &     0.48  &     0.13  &    0.07  &    0  &    0.12  &     0.16 &     0.22  &     0.27 &     0.63  &     0.40 &    0.15  \\ 
\hline \hline
\end{tabular}

        }
    \end{center}
\end{table} 


\section{Summary}
The relative position of classes in the variable-space of a multiple-class classification problem is usually complicated.
Hence, some complicated modeling procedures are preferred for their classification performance.
Using a voting scheme via building many binary classifiers is another choice.  
However, this kind of approach usually aims at high prediction power, and sacrifices the ``model interpretation'' ability, 
which is important from many classification application perspectives. 
If we only aim our classification goal at classification performance, such as accuracy, specificity, sensitivity/precision, then finding only a set of variables for all classes, via using a complicated model, could be an easy way out.  
Nevertheless, we will have less power about model interpretation when the relations among classes are complicated,
and this goal is even more difficult to be achieved when data sets have lengthy variables.
Thus, when the model-interpretation ability is indispensable to an application, deciding class-specific variable sets to clarify the classification rules of models is important in addition to the conventional classification performance measures.
In this study, we adopt NPKDC-vd for multiple-class classification, and shows that the error rate of this method approaches to that of the Bayes rule, asymptotically.
Moreover, through local bandwidths of density estimation, we integrate such a posterior density estimation procedure with a variable selection scheme  such that we are able to identify the individual high impact variable set for each class,  such that the proposed classification rule are easier to be understood,
and we can have a better description of the character of each class.
In the numerical examples, we see that the proposed method has a satisfactory classification performance with only a limited increase of computational burden, and  the ``interpretation'' of each class is largely improved, especially for the examples with complicated relative positions among classes. 
Because we can separately calculate the local bandwidths within each class, the computational efficiency can be further improved via parallel computing methods, and can be easily implemented via machines with a multi-core CPU. 
Moreover, because using the bandwidth to select the relevant variables is a promising approach, 
we suggest a two-step approach such that we can first filter out the irrelevant variables to reduce variable size in advance, 
then apply the proposed algorithm with the rest variables.  
This approach can be useful for problems with a small sample size but a large number of variables.  
Since there will need different considerations for filtering out variables in the first stage, 
and will go beyond the scope of the current project. 
Thus, we will study this problem in our future studies, and will report its results elsewhere.


\end{document}